\documentclass{article}

\PassOptionsToPackage{numbers,sort&compress}{natbib}
\usepackage[preprint]{neurips_2026}

\makeatletter
\renewcommand{\@noticestring}{}
\let\rawinput\@@input
\makeatother

\usepackage[utf8]{inputenc}
\usepackage[T1]{fontenc}
\usepackage{hyperref}
\usepackage{booktabs}
\usepackage{graphicx}
\usepackage{xcolor}
\usepackage{amsmath}
\usepackage{wrapfig}
\usepackage{caption}
\usepackage{placeins}

\usepackage{etoolbox}
\IfFileExists{generated/macros.tex}{\newcommand{\RNScenes}{150}
\newcommand{\RMaskRef}{0.479}
\newcommand{\RMaskRun}{0.517}
\newcommand{\RMaskDiff}{+0.038}
\newcommand{\RMaskCI}{[-0.018,+0.093]}
\newcommand{\RMaskWL}{37/26}
\newcommand{\RMaskP}{0.21}

\newcommand{\RAebRun}{0.599}
\newcommand{\RAebDiff}{+0.082}
\newcommand{\RAebCI}{[+0.023,+0.143]}
\newcommand{\RAebWL}{41/28}
\newcommand{\RAebP}{0.15}

\newcommand{\RAebFiveDiff}{-0.009}
\newcommand{\RAebFiveCI}{[-0.065,+0.044]}
\newcommand{\RAebFiveWL}{34/34}
\newcommand{\RAebFiveP}{1.00}

\newcommand{\RMaskColl}{34.0}
\newcommand{\RAebColl}{11.0}

\newcommand{\RBaseFrontRate}{29.9}
\newcommand{\RBaseFrontN}{90/301}
\newcommand{\RBaseAccel}{42}
\newcommand{\RBaseNoSlow}{68}
\newcommand{\RMaskFrontRate}{25.7}
\newcommand{\RMaskFrontN}{77/300}
\newcommand{\RMaskAccel}{30}
\newcommand{\RMaskNoSlow}{64}
}{}

\title{GatedVLA with Camera-Based Emergency Braking: AlpaSim E2E Closed Loop Challenge Technical Report}

\author{\small\begin{tabular}[t]{@{}c@{\hspace{10pt}}c@{\hspace{10pt}}c@{}}
  Kaneyoshi Hiratsuka & Naoki Takada & Makoto Kawano \\
  \mdseries Kyoto University & \mdseries Tokyo University of Agriculture and Technology & \mdseries The University of Tokyo
\end{tabular}}

\begin{document}
\maketitle

\begin{abstract}
This report presents GatedVLA, our entry for the Physical AI AV (PAI-AV) track. Our learned vision-language-action (VLA) policy pairs the driving-pretrained Qwen-Drive-1.0-4B vision-language model~\cite{qwendrive} with a layer-wise flow-matching action expert that outputs a 2.5\,s horizon of acceleration and curvature controls, augmented with a soft visual mask applied to the expert's cross-attention. Compared to an identically trained baseline, the masked policy achieves comparable overall open-loop error while reducing error on left turns. In local closed-loop evaluation across the same \RNScenes{} scenes (two rollouts per scene), the visual mask increases the mean scene score from \RMaskRef{} to \RMaskRun{} (paired difference $\RMaskDiff$, 95\% bootstrap CI $\RMaskCI$, wins/losses \RMaskWL, sign test $p=\RMaskP$). Incorporating a rule-based automatic emergency braking (AEB) module at deployment further increases the mean score to \RAebRun{} ($\RAebDiff$, CI $\RAebCI$) and reduces the mean number of scenes with at-fault collisions from \RMaskColl{} to \RAebColl{}.
\end{abstract}

\section{Introduction}
Even minor planning errors alter the subsequent observations encountered by a driving policy---a compounding effect that only closed-loop evaluation can reveal. The AlpaSim challenge evaluates policies in closed loop on held-out scenes~\cite{alpasim}, where a scene receives a score of zero if the ego vehicle causes an at-fault collision, drives off-road, or exits the recorded driving corridor. In our preliminary experiments, at-fault collisions were the primary source of score loss, predominantly involving slower or stationary lead vehicles when the planned trajectory failed to initiate braking. We therefore introduce (1) a learned visual mask trained to selectively retain image evidence relevant to navigation intent for the action expert, and (2) a detector-based emergency braking module that intervenes when the learned policy fails to decelerate.

\section{Related work}
\textbf{Vision-language-action driving.} Qwen-Drive-1.0~\cite{qwendrive} establishes a vision-language foundation model for autonomous driving paired with a planning expert. Alpamayo-R1~\cite{alpamayo} couples a reasoning model with a diffusion-based trajectory decoder, while DriveMoE~\cite{drivemoe} introduces expert routing to Drive-$\pi_0$, an adaptation of $\pi_0$~\cite{pi0}. We train a $\pi_0$-style flow-matching expert on top of the Qwen-Drive backbone.
\textbf{Limiting visual evidence.} S2~\cite{seeless} learns per-token soft masks under a visual evidence budget alongside an ungated pathway. We adopt a similar gating head and budget formulation, but apply the mask directly to the action expert's cross-attention logits, enforce an exact budget by construction, and train the mask via self-distillation (Section~\ref{sec:policy}).
\textbf{Copycat behavior.} Imitation learning policies can suffer from causal confusion by over-relying on their own past motion~\cite{causal}; ChauffeurNet~\cite{chauffeurnet} mitigates this with history dropout during training. We explored both history dropout and temporal ensembling~\cite{act}, but observed no clear gains (Appendix~\ref{app:others}).

\section{Method summary}
\begin{figure}[h]
\centering
\includegraphics[width=0.72\linewidth,trim=7.5 10 164 148,clip]{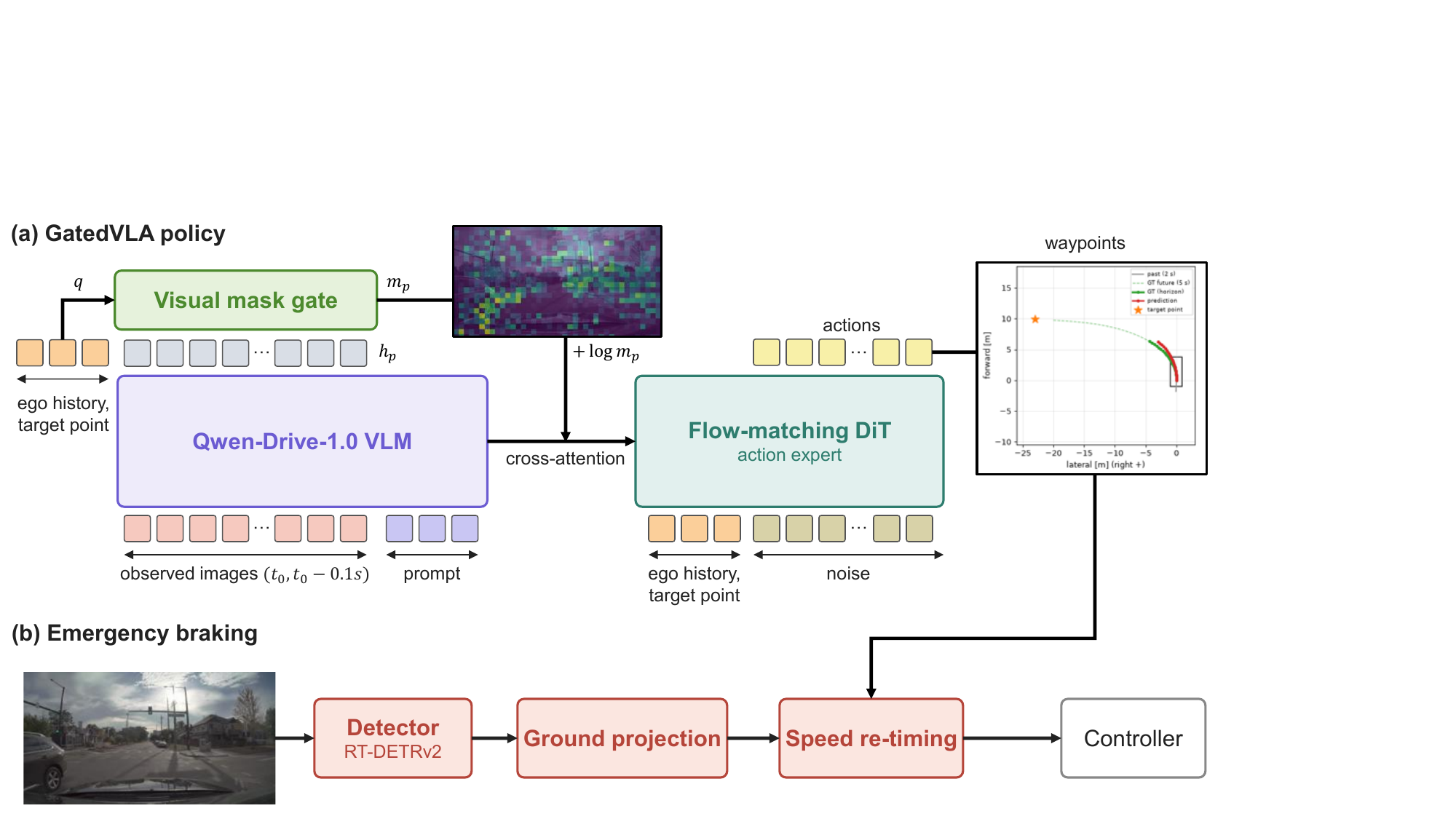}
\caption{System overview. (a) GatedVLA policy, trained end-to-end: the visual mask $m_p$ biases the expert's cross-attention to the VLM's image tokens. (b) Pretrained detector and rule-based AEB, integrated only at deployment: they re-time the planned waypoints along their path.}
\label{fig:system}
\end{figure}

Our driving system (Fig.~\ref{fig:system}) comprises an end-to-end learned policy that replans every 0.5\,s and a detector-based AEB module that exclusively decelerates the planned trajectory along its predicted path. We evaluated and submitted three deployment variants derived from a single trained checkpoint: AEB at 10\,Hz, AEB at 5\,Hz, and the policy without AEB. Our implementation builds upon starVLA~\cite{starvla} and py123d\_garage~\cite{garage,py123d}, and all evaluations are conducted in AlpaSim~\cite{alpasim}.

\section{Policy description}
\label{sec:policy}
\textbf{Inputs.} Two front wide-angle camera frames (at $t_0$ and $t_0{-}0.1$\,s) along with the fixed text prompt ``predict trajectory'' are fed into the VLM. Ego-state history and navigation conditioning are provided exclusively to the action expert: five 10\,Hz historical steps of $[x, y, \mathrm{heading}, \mathrm{speed}, a_\mathrm{lon}]$ concatenated at each step with a target point $(x_\mathrm{tp}, y_\mathrm{tp})$ and a binary target-presence flag. During training, the target point is the first future ego position at a Euclidean distance of at least 25\,m; at deployment, it is the first such point along AlpaSim's map route (Appendix~\ref{app:driver}).

\textbf{Expert and outputs.} The action expert follows a $\pi_0$-style layer-wise architecture~\cite{pi0,drivemoe}: a 32-block Diffusion Transformer (DiT)~\cite{dit} alternates self-attention and cross-attention layers, where each cross-attention block attends to a corresponding VLM layer via a learned linear projection. The expert is trained via conditional flow matching~\cite{flowmatching} with a velocity MSE objective and generates trajectories using 10 Euler integration steps at inference. It predicts 25 steps of $[a_\mathrm{lon}, \mathrm{curvature}]$ controls at 10\,Hz, utilizing the kinematic unicycle action space adapted from NVIDIA Alpamayo~1.5~\cite{alpamayocode,alpamayo}. Integrating these controls from the current ego speed yields the output poses.

\textbf{Baseline.} The baseline shares the same architecture and training protocol but omits the visual mask and Paths B and C (below); as in Path A, the target point is dropped with probability $p=0.5$ per sample.

\textbf{Visual mask.} A gating module takes each visual token's detached final-layer VLM hidden state $h_p$ alongside a conditioning query $q$ (which embeds the scaled target point and ego state) and computes a logit score from $[h_p, q, h_p\odot q]$, following the visual evidence budget formulation of S2~\cite{seeless}. The scores are standardized across tokens within each frame to obtain $z_p$, yielding soft mask values $m_p=\sigma(b+2z_p)$, where the scalar bias $b$ is solved via bisection such that the per-frame mean of $m_p$ equals the target budget $\rho=0.2$ by construction, eliminating the need for an auxiliary budget loss. We add $\log m_p$ directly to the expert's cross-attention logits while leaving text tokens and internal VLM representations unmasked. Each training step executes three expert passes sharing a single VLM forward pass, sampled noise, and flow time step:
\vspace{-5pt}
\begin{itemize}\setlength\itemsep{0pt}\setlength\parskip{0pt}\setlength\parsep{0pt}
\item \textbf{Path A} (ungated, target point dropped with $p=0.5$): minimizes the standard flow-matching loss and updates both the VLM and DiT.
\item \textbf{Path B} (stop-gradient mask, target point provided): minimizes $\lVert v_B-\mathrm{sg}(v_A)\rVert^2$ to update the DiT; this serves as the inference pathway.
\item \textbf{Path C} (active mask, target point omitted from the DiT state but provided to the gate): minimizes $\lVert v_C-\mathrm{sg}(v_A)\rVert^2$ and updates only the gating module.
\end{itemize}
\vspace{-5pt}
Paths B and C are distilled only on samples where the target point is retained in Path A; consequently, Path C forces the gating module to preserve visual tokens that encode the navigation intent.

\textbf{Emergency brake.} An RT-DETRv2-R18 detector~\cite{rtdetrv2,rtdetr} pretrained on COCO~\cite{coco} processes the current front-camera frame and retains boxes of seven traffic classes with confidence scores exceeding 0.5. The central 40\% of each bounding box's bottom edge is projected onto the ego rig's flat ground plane using the nominal $f$-$\theta$ camera calibration. Among detected obstacles within a 1.3\,m lateral corridor of the planned path, the closest obstacle defines the arc-length gap $d$ measured from the front bumper along the path; the closing speed $c$ is estimated via linear regression over the last 1\,s of $d$ (Table~\ref{tab:app_deploy}). With braking margin $g$ and required deceleration $a_\mathrm{req}$,
\begin{equation}
g = d - 2.5\,\mathrm{m} - 0.4\,\mathrm{s}\cdot c, \qquad a_\mathrm{req} = \frac{c^2}{2\max(g,\,0.1\,\mathrm{m})},
\end{equation}
the planned speed profile is capped by a linear deceleration ramp from the current speed $v$ down to the estimated lead speed $\max(v-c,0)$ at a rate of $\mathrm{clip}(1.2\,a_\mathrm{req},3,8)$\,m/s$^2$ whenever $g\le 0$ or $a_\mathrm{req}>3$\,m/s$^2$. Under a milder time-to-collision condition ($d/c<3$\,s), the planned speed is constrained not to exceed the current ego speed. Because the trajectory is re-timed along its own path from the ego vehicle's projection, a braked ego will not accelerate to catch up with a stale plan. Upon AEB release, the speed is held for one control tick before an immediate policy replan.

\section{Resources and training}
\textbf{Data.} Models are trained on the NVIDIA PhysicalAI-Autonomous-Vehicles dataset~\cite{paiav} (NVIDIA Autonomous Vehicle Dataset License Agreement), using only the front wide-angle camera streams and egomotion trajectories from the training split. No LiDAR, radar, or machine-generated perception labels were used.

\textbf{Pretrained weights.} We initialize the VLM from Qwen-Drive-1.0-4B~\cite{qwendrive} (Apache-2.0) and use off-the-shelf RT-DETRv2-R18 weights (\texttt{PekingU/rtdetr\_v2\_r18vd}, Apache-2.0) for detection. \textbf{Private resources:} None.

\textbf{Training.} Both the baseline (W\&B run \texttt{ccb412c9}) and masked policy (\texttt{03bdfc50}) were trained from the same code commit with random seed 42 for 12 epochs (38{,}400 steps) across 12 Intel Aurora XPU tiles. In the spirit of linear probing then fine-tuning (LP-FT)~\cite{lpft}, the VLM backbone remained frozen during the first epoch. Full hyperparameter settings and training curves are provided in the appendix.

\begin{wraptable}{r}{0.56\linewidth}
\centering\footnotesize
\caption{Open-loop prediction error at step 38{,}400 (baseline / mask). ADE and FDE in meters over 2.5\,s; miss: any waypoint error $>2$\,m.}
\label{tab:openloop}
\begin{tabular}{@{}lccc@{}}
\toprule
 & ADE & FDE & Miss rate \\
\midrule
All & 0.337 / 0.341 & 1.029 / 1.023 & 0.125 / 0.125 \\
Left & 0.405 / 0.376 & 1.213 / 1.100 & 0.200 / 0.150 \\
Right & 0.367 / 0.360 & 1.136 / 1.089 & 0.087 / 0.100 \\
Straight & 0.239 / 0.287 & 0.737 / 0.879 & 0.087 / 0.125 \\
\bottomrule
\end{tabular}
\end{wraptable}
\textbf{Open-loop results.} Table~\ref{tab:openloop} reports open-loop trajectory prediction performance on a fixed validation subset of 240 samples (80 left turns, 80 right turns, and 80 straight segments). Overall open-loop error is comparable between the two models: the visual mask improves performance on left turns (FDE of 1.100 vs.\ 1.213; miss rate of 0.150 vs.\ 0.200) and yields a slightly lower FDE on right turns, at the cost of higher error on straight segments. Ablating visual inputs degrades FDE substantially more than omitting the navigation target point (masked policy: 1.564 with black frames vs.\ 1.135 with no target and no mask), confirming that visual evidence strongly informs planning in both models (baseline with black frames: FDE 1.408). Appendix~\ref{app:open} gives all controls.

\textbf{Closed-loop results.} Table~\ref{tab:closedloop} summarizes local closed-loop evaluation across the same \RNScenes{} scenes sampled from the 441-scene curated validation split using the \texttt{dev} challenge preset (40\,m route offset). A rollout without such a failure scores $\min(\mathrm{progress}/0.8,1)$. The first and second rollouts ran on different hosts (NVIDIA Blackwell and Ada), so differences between them reflect both hardware and run-to-run variance. Averaging scores across the two rollouts per scene: mask vs.\ baseline achieves a mean difference of $\RMaskDiff$ (95\% bootstrap CI $\RMaskCI$, wins/losses \RMaskWL, sign test $p=\RMaskP$); AEB at 10\,Hz vs.\ mask alone yields $\RAebDiff$ (CI $\RAebCI$, \RAebWL, $p=\RAebP$); and AEB at 5\,Hz vs.\ 10\,Hz yields $\RAebFiveDiff$ (CI $\RAebFiveCI$, \RAebFiveWL, $p=\RAebFiveP$). Appendix~\ref{app:closed} gives all pairwise comparisons.

\begin{table}[h]
\centering\footnotesize
\caption{Local closed-loop results across the same \RNScenes{} scenes. Failure counts are numbers of affected scenes; rear-end collisions by other agents are not at-fault and merely halt progress accrual. Mean Drive latency includes driver video recording.}
\label{tab:closedloop}
\begin{tabular}{llcccccc}
\toprule
Configuration & Host & Score & At-fault & Offroad & Corridor exit & Rear-ended & Drive (s) \\
\midrule
Baseline, 1st & Blackwell & 0.476 & 42 & 25 & 8 & 12 & 0.133 \\
Baseline, 2nd & Ada & 0.481 & 41 & 19 & 15 & 9 & 0.066 \\
Mask, 1st & Blackwell & 0.499 & 35 & 22 & 12 & 15 & 0.110 \\
Mask, 2nd & Ada & 0.534 & 33 & 23 & 8 & 15 & 0.146 \\
Mask + AEB 10\,Hz, 1st & Blackwell & 0.607 & 12 & 27 & 11 & 25 & 0.201 \\
Mask + AEB 10\,Hz, 2nd & Ada & 0.591 & 10 & 31 & 13 & 23 & 0.085 \\
Mask + AEB 5\,Hz & Ada & 0.590 & 13 & 27 & 10 & 28 & 0.264 \\
\bottomrule
\end{tabular}
\end{table}

\section{Runtime and reproducibility}
\textbf{Serving.} The submission container image is self-contained, operates within the challenge sandbox without external network access, and launches without arguments. Between 0.5\,s replans, it serves poses from the cached plan at 10\,Hz.

\textbf{Latency and memory.} The mean \texttt{Drive} latencies reported in Table~\ref{tab:closedloop} were recorded locally with driver video recording enabled and up to 12 concurrent rollouts sharing a single driver on multi-tenant hosts, so they are only indicative. In a standalone run with a single rollout per replica and video recording disabled (on a shared NVIDIA RTX 6000 Ada GPU), the mean \texttt{Drive} RPC latency was 0.10--0.11\,s for both AEB variants (Appendix~\ref{app:runtime}). The RT-DETRv2 detector requires approximately 0.2\,GiB of VRAM in addition to the policy's 11.6\,GiB footprint.

\textbf{Reproducibility.} The deployed inference configuration matches our logged W\&B training runs. Source code: \url{https://github.com/Azuma413/GatedVLA} (branch \texttt{main}, commit \texttt{80ab247}).

\section{Limitations and safety considerations}
\label{sec:limits}
\textbf{Limited generalization evidence for AEB.} The AEB decision thresholds were tuned offline by replaying cached detections on the same 150 scenes evaluated in Table~\ref{tab:closedloop}, so the observed gains may be optimistic for held-out private test scenes.

\textbf{Remaining collisions.} Of the 12 remaining at-fault collisions in the first 10\,Hz AEB rollout (manual inspection, Appendix~\ref{app:closed}), 6 involved obstacles offset by approximately 2\,m laterally that entered the 1.3\,m detection corridor too late or not at all, a trade-off from narrowing the corridor to suppress false-positive braking; the rest involved agents from the side or rear-left (2), low-speed contacts where distance estimates stagnated (2), or braking too late or not to a complete stop (2).

\textbf{Side effects of emergency braking.} In the first rollout set, AEB increases the number of rear-ended scenes from 15 to 25. While these are not at-fault in the simulator, abrupt braking poses a genuine safety hazard in real-world traffic; moreover, 3 previously full-score scenes incurred new at-fault collisions under AEB.

\textbf{Dominance of lateral failures.} In the first AEB rollout, off-road departures and lateral corridor exits account for 0.253 of the 0.393 total score loss---failure modes that longitudinal AEB cannot mitigate. Off-road violations occurred at a median deviation of only 0.67\,m from the map route (comparable to normal deviations in successful rollouts), so we opted against deploying a route-distance guard.

\textbf{Suspected reliance on ego motion.} Without AEB, first-incident front collisions occur in \RMaskFrontRate\% of masked policy rollouts (\RMaskFrontN) vs.\ \RBaseFrontRate\% for the baseline (\RBaseFrontN). The visual mask appears to reduce this failure mode, but the final plan commanded no deceleration in \RMaskNoSlow\% vs.\ \RBaseNoSlow\% of these collisions (Table~\ref{tab:app_kin}).

\bibliographystyle{abbrvnat}
\bibliography{references}

\appendix
\captionsetup{font=footnotesize,skip=4pt}
\setlength{\intextsep}{6pt}
\setlength{\columnsep}{12pt}

\FloatBarrier
\section{Full configuration}
\label{app:config}
Table~\ref{tab:app_cfg} lists the training configuration and Table~\ref{tab:app_deploy} the deployed driver and AEB configuration.

\begin{table}[!htb]
\centering\scriptsize
\caption{Training configuration (identical for both runs unless otherwise stated). Source: W\&B configurations of \texttt{ccb412c9} and \texttt{03bdfc50}; the model bundles deployed in the container images carry identical values.}
\label{tab:app_cfg}
\begin{tabular}{@{}lp{0.74\linewidth}@{}}
\toprule
Item & Value \\
\midrule
Epochs / steps & 12 / 38{,}400 (3{,}200 steps per epoch) \\
Hardware & 1 Aurora node, 12 XPU tiles, one rank per tile (batch size 4 per tile) \\
Precision & bf16; gradient checkpointing enabled; max gradient norm 1 \\
Optimizer & AdamW, betas $(0.9, 0.95)$, $\epsilon=10^{-8}$, weight decay 0 \\
Learning rates & VLM $10^{-6}$; expert $5\times10^{-5}$; gate $10^{-4}$ (mask run) \\
Schedule & 200 warmup steps, cosine decay to $10^{-8}$; VLM frozen in epoch 1, then warmed up and decayed over the rest \\
Repeated diffusion steps & 4 noise/time draws per image \\
Prompt & ``predict trajectory'' (no target or speed in the prompt) \\
Images & Front wide-angle camera; pixel budgets 921{,}600 ($t_0$) and 174{,}080 ($t_0-0.1$\,s); 1 token per $32\times32$ pixels \\
VLM & \texttt{Qwen/Qwen-Drive-1.0-4B}: 32 layers, hidden dim 2560, SDPA attention \\
Expert & LayerwiseFM DiT: 32 blocks interleaving cross- and self-attention, width 1024, 16 heads of dim 64, dropout 0.2, AdaNorm \\
State & Horizon 5, dim 8: $x, y, \mathrm{heading}, \mathrm{speed}, a_\mathrm{lon}, x_\mathrm{tp}, y_\mathrm{tp}$, flag; per-dim q01/q99 scaled to $[-1,1]$ \\
Action & Horizon 25, dim 2: acceleration, curvature; per-dim z-score \\
Flow matching & Beta(1.5, 1) time, $s=0.999$; velocity MSE; 10 Euler steps at inference; 1000 time buckets \\
Target dropout & $p=0.5$ per sample (Path A in the mask run); target columns zeroed, flag set to 0 \\
Eval & Every 250 steps on 240 fixed validation samples; open-loop \\
\midrule
Mask (mask run only) & Scheme \texttt{three\_path}; budgeted parameterization; $\rho=0.2$; contrast 2; head hidden dim 256 \\
Gate query & Target point $/25$ and flattened normalized ego state without target columns \\
Gate input & Final-layer VLM hidden states of visual tokens, detached, standardized per channel; RMSNorm \\
\bottomrule
\end{tabular}
\end{table}

\begin{table}[!htb]
\centering\scriptsize
\caption{Deployment configuration. Replanning, CUDA graph, and detector settings correspond to default parameters or pre-configured overrides baked into the submitted container images.}
\label{tab:app_deploy}
\begin{tabular}{@{}lp{0.70\linewidth}@{}}
\toprule
Item & Value \\
\midrule
Replanning & Every 5th 100\,ms interval (0.5\,s); plan cached in the local frame and re-sliced at each tick \\
Temporal ensembling; action samples & Off (1 plan; 1 sample) across all three submissions \\
Linear-attention kernels & \texttt{auto}: flash-linear-attention Triton kernels, falling back to the reference implementation \\
Expert denoising & Replayed from a captured CUDA graph \\
Target point & 25\,m along the route; route resampled at 0.1\,m and extended 2\,m past the farthest target \\
Controller gains & Official defaults (no \texttt{--controller-gains} override) \\
\midrule
Detector & RT-DETRv2-R18, COCO, $640\times640$ on the front wide-angle frame, score $>0.5$, no NMS (DETR queries) \\
Classes & person, bicycle, car, motorcycle, bus, train, truck \\
Ground projection & 7 points along the central 40\% of the box bottom edge; flat ground $z=0$ in the rig frame; range $<80$\,m \\
Corridor & 1.3\,m half-width around the planned path; path linearly extrapolated to 80\,m \\
Front overhang & 4.05\,m (rear axle to front bumper); detections behind $-1$\,m relative to the bumper ignored \\
Track & 1\,s window; at least 6 samples (10\,Hz) or 3 (5\,Hz); reset on a jump above 4\,m or missed detection \\
Stop gap; latency & 2.5\,m; 0.4\,s \\
Brake trigger; deceleration & $a_\mathrm{req}>3$\,m/s$^2$ or $g\le0$; $\mathrm{clip}(1.2\,a_\mathrm{req},3,8)$\,m/s$^2$ \\
No-acceleration condition & $d/c<3$\,s \\
Re-timed plan grid & 0.1\,s \\
Detection rate & Every front frame (10\,Hz variant); at least 0.2\,s apart (5\,Hz variant) \\
\bottomrule
\end{tabular}
\end{table}

\FloatBarrier
\section{Data and action representation}
\label{app:data}
\textbf{Sampling.} We use all clips containing valid egomotion and front-camera streams from the dataset's training split. Each dataset item corresponds to a single clip with the anchor timestamp $t_0$ sampled uniformly from its valid range, ensuring that each epoch visits every clip once at a random timestamp. Only the densely sampled portion of the egomotion trajectory (spanning approximately 20\,s around the camera window) is loaded. The two visual inputs are the front wide-angle frames closest to $t_0$ and $t_0-0.1$\,s.

\textbf{State.} At each historical step $t \in \{t_0-0.4\,\mathrm{s},\dots,t_0\}$, the ego state consists of $[x, y, \mathrm{heading}_\mathrm{rel}, \mathrm{speed}, a_\mathrm{lon}]$ expressed in the ego coordinate frame at $t_0$. All of these quantities are directly provided by AlpaSim at deployment. The target point, also expressed in the ego frame at $t_0$, is repeated across all historical steps and followed by a binary presence flag.

\textbf{Target point.} The target point is defined as the first future ego position located at a Euclidean distance of at least 25\,m from the current position, or the final recorded position of the clip if the trajectory terminates earlier. In the training run for our deployed model, this fallback endpoint case was retained (a subsequent experiment in Appendix~\ref{app:others} omits it, because a clip endpoint implicitly indicates where the ego vehicle stops---information that a map-based route target never provides at deployment).

\textbf{Action.} The action representation consists of 25 per-step [acceleration, curvature] pairs starting from $t_0$, extracted via Alpamayo~1.5's regularized least-squares fitting procedure (which jointly smooths heading, velocity, acceleration, and curvature, clamping curvature to 0 at speeds below 0.6\,m/s) and integrated via the trapezoidal rule from the current speed $v_0$ to yield 2D poses at $t_0+0.1\,\mathrm{s},\dots,t_0+2.5$\,s.

\textbf{Statistics and validation.} Feature normalization statistics are computed from 6{,}000 training clips $\times$ 8 time steps. The fixed validation set comprises 240 samples stratified into 80 samples per maneuver category, classified by the cumulative heading change over the subsequent 3\,s using a $\pm15^\circ$ threshold.

\FloatBarrier
\section{Mask details and diagnostics}
\label{app:mask}
\textbf{Budgeted mask.} Given raw token scores $\ell_p$ within a single frame, we standardize them as $z_p=(\ell_p-\bar\ell)/(\mathrm{std}(\ell)+10^{-6})$ and compute soft mask values $m_p=\sigma(b+\alpha z_p)$ with $\alpha=2$, where 40 bisection iterations over $b\in[-30,30]$ enforce $\mathrm{mean}_p\,m_p=\rho$. The bisection solver itself is not differentiated directly. Instead, because the constraint $\sum_p\sigma(b+\alpha z_p)=N\rho$ defines $b(z)$ implicitly with $\partial b/\partial z_q=-\alpha w_q$ and $w=\sigma'/\sum\sigma'$, we add a zero-valued surrogate term $-\alpha\sum_q w_q(z_q-\mathrm{sg}[z_q])$ to supply this exact analytical gradient during backpropagation. By construction, a uniform mask is precluded and the dispersion of $m$ is fixed by $\alpha$, preventing the mask from collapsing to an all-keep state or degenerating into a binary mask.

\textbf{Where it acts.} In every cross-attention block of the action expert, the attention logits are biased by adding $\log m_p$ to visual tokens, $0$ to text tokens, and $-\infty$ to padding tokens. The VLM itself remains unmasked. A single shared gating head predicts masks for both input frames, enforcing the budget constraint independently per frame. Unlike S2~\cite{seeless}, which multiplies token embeddings by a floored mask inside the action executor, the mask here acts only through these attention logits.

\begin{wraptable}{r}{0.5\linewidth}
\centering\scriptsize
\caption{Mask diagnostics across the validation set (current frame) at step 38{,}400.}
\label{tab:app_diag}
\begin{tabular}{@{}p{0.8\linewidth}c@{}}
\toprule
Diagnostic & Value \\
\midrule
Mean keep, history frame & 0.200 \\
Mean keep, current frame & 0.200 \\
Spearman rank correlation between masks of two samples (1 = identical fixed pattern) & 0.055 \\
Mask variance explained by the mean row pattern & 0.024 \\
Mask variance explained by the mean column pattern & 0.003 \\
Spearman rank correlation between a sample's mask and its DiT attention & 0.827 \\
DiT cross-attention mass on text tokens & 0.261 \\
Keep on image left minus right half, left turns & 0.002 \\
Same, right turns & 0.025 \\
Same, straight & 0.011 \\
\bottomrule
\end{tabular}
\end{wraptable}
\textbf{Mask diagnostics.} At step 38{,}400 the mask depends on the sample: the Spearman rank correlation between the masks of two samples is 0.055, and the mean row and column patterns explain 2.4\% and 0.3\% of its variance. The DiT's cross-attention largely follows the mask (rank correlation 0.827), and 26.1\% of the attention mass falls on text tokens. The mask shows no clear lateral preference by maneuver: keep on the left minus the right image half is 0.002, 0.025, and 0.011 for left turns, right turns, and straight driving (Table~\ref{tab:app_diag}).

\textbf{Gradients.} Path A updates both the VLM and the DiT. Path B uses a detached mask and detached VLM features to update only the DiT. Path C uses detached VLM features, and the trainer applies its gradients exclusively to the gating module. The combined action loss is $L_A+L_B$, whereas $L_C$ is backpropagated in a separate pass. All three pathways share the same sampled noise and flow time step, and losses $L_B$ and $L_C$ are weighted by whether the sample retained the target point in Path A.

\textbf{Controls.} \emph{No mask}: identical inputs, omitting the $\log m$ bias. \emph{Shuffled}: randomly permuting the predicted retention weights $m_p$ across spatial positions within each frame. \emph{Mirrored}: evaluating the gating module with a laterally mirrored target point ($y\to-y$). \emph{No image tokens}: setting retention weights to $0$ on every visual token. \emph{Black frames}: replacing all camera inputs with black images and re-running the VLM without masking. \emph{No target}: setting the target-presence flag to $0$ and zeroing the target columns in the DiT state; when combined with the mask, target information reaches the policy solely through the gating module.

\begin{figure}[!htb]
\centering
\includegraphics[width=0.9\linewidth]{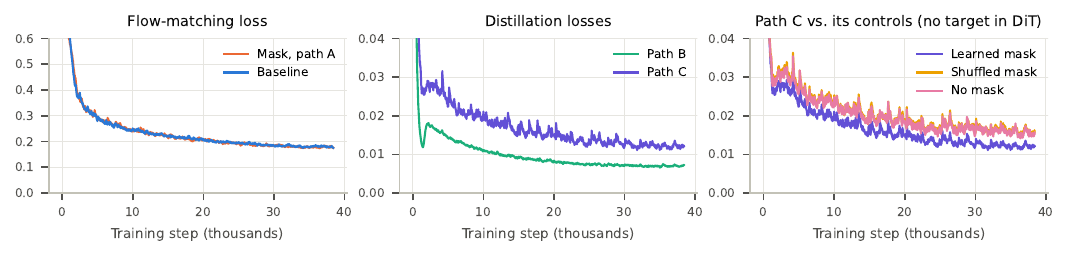}
\caption{Smoothed training losses (EMA factor 0.05). The baseline flow-matching loss closely tracks Path A of the masked run, while Path C with the learned mask remains consistently below its shuffled-mask and unmasked controls.}
\label{fig:curve_losses}
\end{figure}

\begin{wraptable}{r}{0.56\linewidth}
\centering\scriptsize
\caption{Training signals: mean of the first 50 and last 200 logged points (logged every 10 steps).}
\label{tab:app_train}
\begin{tabular}{@{}lcccc@{}}
\toprule
 & \multicolumn{2}{c}{Mask run} & \multicolumn{2}{c}{Baseline} \\
Signal & First & Last & First & Last \\
\midrule
Flow-matching loss (Path A) & 0.9586 & 0.1790 & 0.9424 & 0.1796 \\
Loss, acceleration & -- & -- & 0.9591 & 0.2635 \\
Loss, curvature & -- & -- & 0.9256 & 0.0957 \\
Path B (distillation, DiT) & 0.0521 & 0.0071 & -- & -- \\
Path C (distillation, gate) & 0.0674 & 0.0122 & -- & -- \\
Path C, shuffled mask & 0.0688 & 0.0160 & -- & -- \\
Path C, no mask & 0.0680 & 0.0156 & -- & -- \\
Mean keep, current frame & 0.2000 & 0.2000 & -- & -- \\
Mean keep, history frame & 0.2000 & 0.2000 & -- & -- \\
Std of gate scores & 2.0426 & 3.4981 & -- & -- \\
Fraction with target (Path A) & 0.4987 & 0.5026 & 0.4944 & 0.4997 \\
Gradient norm & 24.555 & 4.1008 & 24.700 & 3.7741 \\
Samples per second & 20.943 & 6.6069 & 28.595 & 6.9585 \\
Step time (s) & 2.2586 & 7.1135 & -- & -- \\
\bottomrule
\end{tabular}
\end{wraptable}
\textbf{Training signals.} The flow-matching loss of Path A falls from 0.96 to 0.18 and ends level with the baseline's (Table~\ref{tab:app_train}). Mean keep stays at 0.200 by construction. Path C ends at 0.0122, below its shuffled-mask (0.0160) and no-mask (0.0156) controls, so the learned mask does make Path C closer to Path A when the DiT receives no target; these controls carry no gradient. The step time rises from 2.3\,s to 7.1\,s and throughput falls accordingly once the VLM is unfrozen after the first epoch. For the baseline, the curvature loss ends well below the acceleration loss (0.096 vs.\ 0.264). Fig.~\ref{fig:curve_losses} shows the same signals over training.

\begin{figure}[!htb]
\centering
\includegraphics[width=\linewidth]{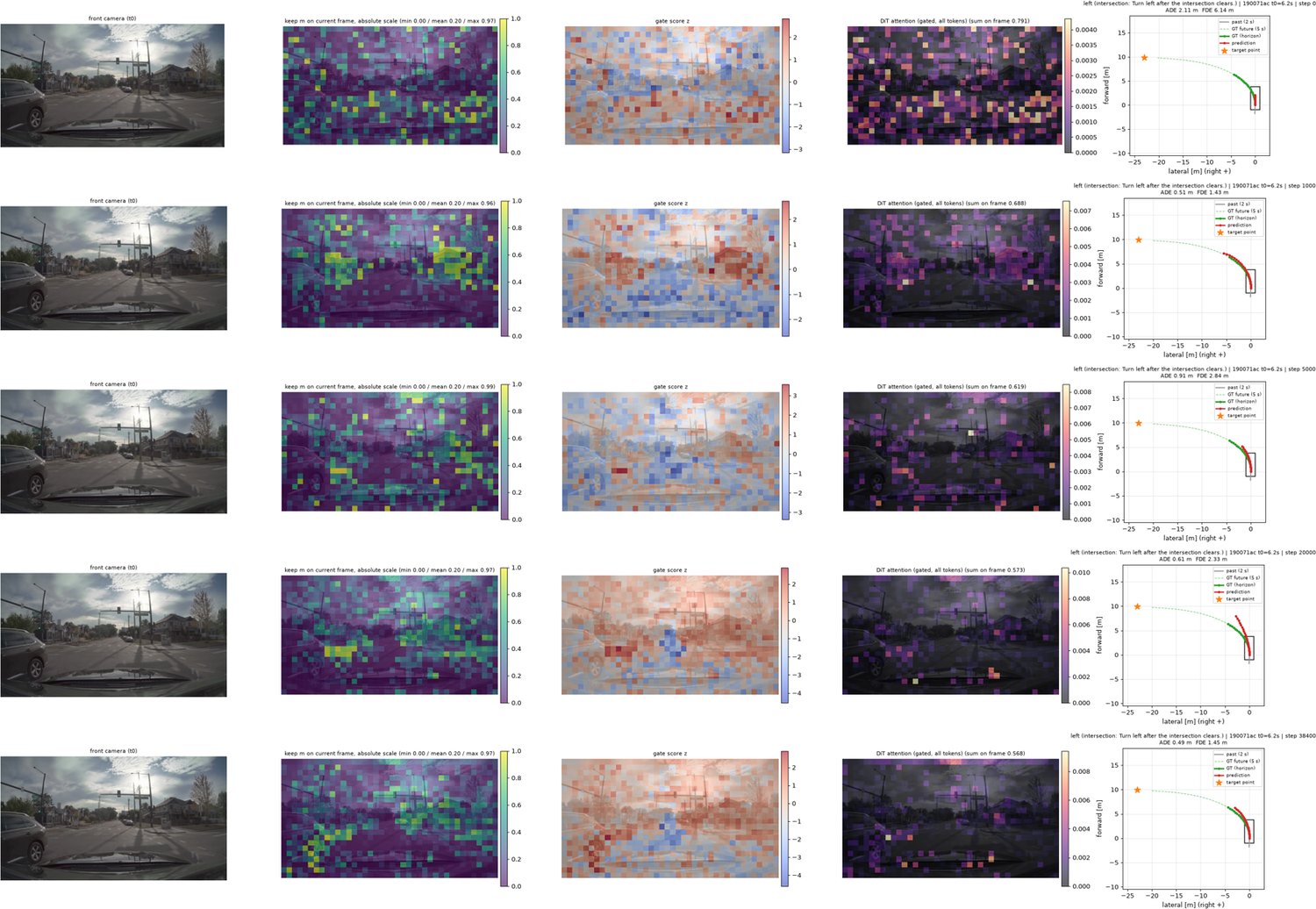}
\caption{Evolution of the visual mask on a left-turn validation sample across training: steps 0, 1{,}000, 5{,}000, 20{,}000, and 38{,}400 (top to bottom). Panels from left to right: current front-camera frame; retention weight $m$; standardized gate score $z$; DiT cross-attention map on the frame; and bird's-eye view showing past trajectory, ground truth, prediction, and target point (lateral axis positive to the right).}
\label{fig:evo_left}
\end{figure}

\begin{figure}[!htb]
\centering
\includegraphics[width=\linewidth]{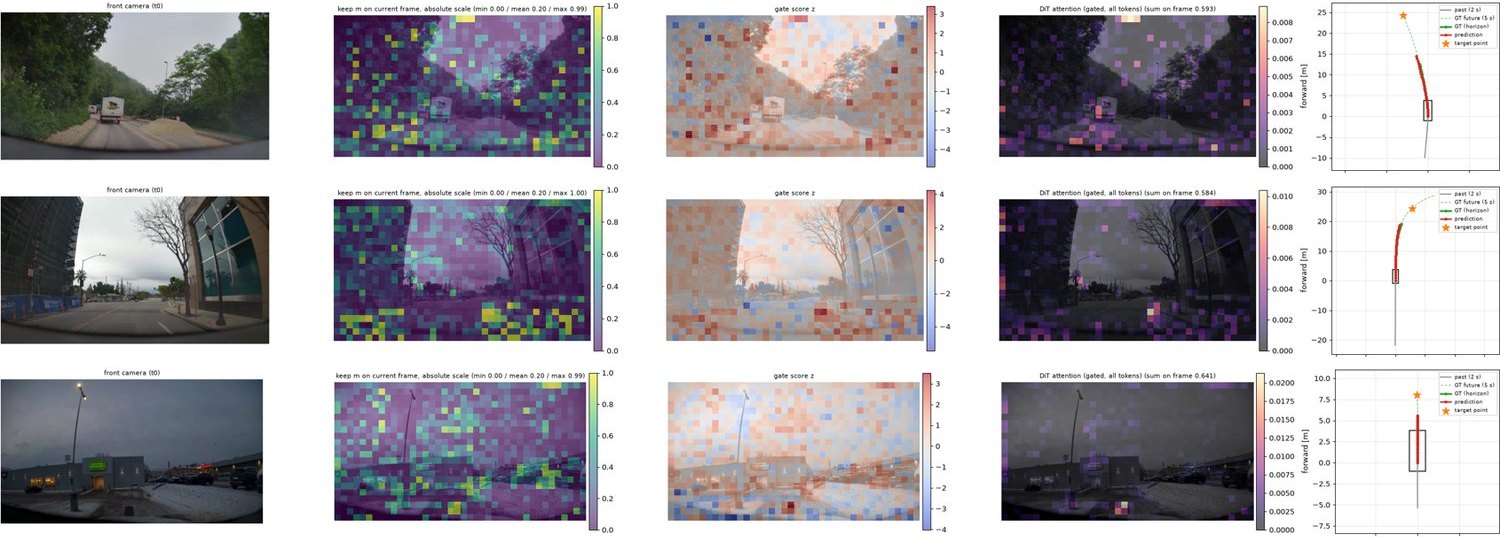}
\caption{Mask-run validation samples at step 38{,}400: a left turn, a right turn, and straight driving (top to bottom), with panels arranged as in Fig.~\ref{fig:evo_left}.}
\label{fig:final}
\end{figure}

\begin{figure}[!htb]
\centering
\includegraphics[width=0.9\linewidth]{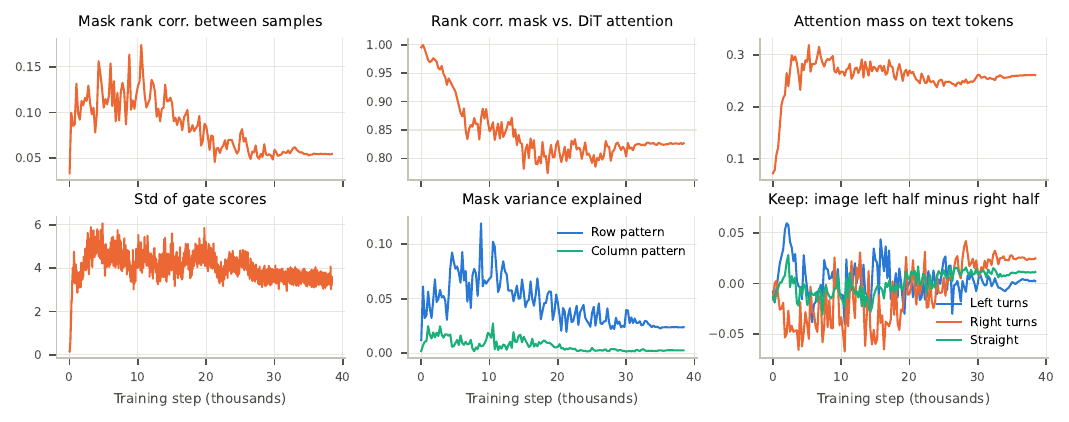}
\caption{Mask diagnostics across training. The rank correlation between the mask and DiT cross-attention decreases from near 1 to approximately 0.83, cross-attention mass on text tokens increases early in training, and mean row/column spatial patterns account for minimal mask variance.}
\label{fig:curve_mask}
\end{figure}

\FloatBarrier
\section{Open-loop results in full}
\label{app:open}
Both models use the final training checkpoint at step 38{,}400. On the fixed 240-sample validation set, the lowest validation error observed during training is slightly lower than that of the final checkpoint: the baseline achieved a minimum ADE of 0.302 at step 32{,}250 (vs.\ 0.337 final), and the masked policy achieved 0.324 at step 34{,}750 (vs.\ 0.341 final). We did not perform checkpoint selection on this validation set. Because each open-loop evaluation draws a single flow-matching trajectory sample per input, these metrics include stochastic sampling variance; no confidence intervals were computed.

\begin{table}[!htb]
\centering\scriptsize
\caption{Open-loop metrics at step 38{,}400 on the 240 fixed validation samples (in meters; miss rate as a fraction). L2 denotes position error at the specified horizon (in seconds); Lat./Lon.\ denote absolute final lateral and longitudinal errors.}
\label{tab:app_open}
\begin{tabular}{llcccccccccc}
\toprule
 & Run & ADE & FDE & L2@0.5 & L2@1.0 & L2@1.5 & L2@2.0 & L2@2.5 & Lat. & Lon. & Miss \\
\midrule
All & Baseline & 0.337 & 1.029 & 0.025 & 0.117 & 0.300 & 0.574 & 0.943 & 0.478 & 0.802 & 0.125 \\
 & Mask & 0.341 & 1.023 & 0.028 & 0.126 & 0.308 & 0.578 & 0.939 & 0.457 & 0.796 & 0.125 \\
\midrule
Left & Baseline & 0.405 & 1.213 & 0.034 & 0.151 & 0.365 & 0.686 & 1.115 & 0.768 & 0.790 & 0.200 \\
 & Mask & 0.376 & 1.100 & 0.034 & 0.146 & 0.347 & 0.637 & 1.014 & 0.667 & 0.712 & 0.150 \\
\midrule
Right & Baseline & 0.367 & 1.136 & 0.025 & 0.117 & 0.320 & 0.633 & 1.041 & 0.516 & 0.931 & 0.087 \\
 & Mask & 0.360 & 1.089 & 0.031 & 0.136 & 0.321 & 0.606 & 0.997 & 0.526 & 0.867 & 0.100 \\
\midrule
Straight & Baseline & 0.239 & 0.737 & 0.017 & 0.083 & 0.215 & 0.404 & 0.672 & 0.149 & 0.686 & 0.087 \\
 & Mask & 0.287 & 0.879 & 0.020 & 0.097 & 0.255 & 0.490 & 0.806 & 0.179 & 0.810 & 0.125 \\
\bottomrule
\end{tabular}
\end{table}

\begin{table}[!htb]
\centering\tiny
\caption{Diagnostic controls for the masked policy at step 38{,}400, sharing the sampled noise of the main prediction. Each three-column group reports ADE, FDE, and miss rate.}
\label{tab:app_ctrl}
\begin{tabular}{lcccccccccccc}
\toprule
 & \multicolumn{3}{c}{All} & \multicolumn{3}{c}{Left} & \multicolumn{3}{c}{Right} & \multicolumn{3}{c}{Straight} \\
\midrule
Inference path & 0.341 & 1.023 & 0.125 & 0.376 & 1.100 & 0.150 & 0.360 & 1.089 & 0.100 & 0.287 & 0.879 & 0.125 \\
\midrule
No mask & 0.339 & 1.016 & 0.129 & 0.376 & 1.096 & 0.138 & 0.358 & 1.086 & 0.125 & 0.282 & 0.866 & 0.125 \\
Shuffled mask & 0.344 & 1.031 & 0.133 & 0.382 & 1.115 & 0.150 & 0.363 & 1.099 & 0.125 & 0.287 & 0.879 & 0.125 \\
Mirrored target in gate & 0.348 & 1.049 & 0.129 & 0.392 & 1.154 & 0.175 & 0.363 & 1.114 & 0.087 & 0.288 & 0.881 & 0.125 \\
No target in DiT, no mask & 0.367 & 1.135 & 0.167 & 0.395 & 1.217 & 0.188 & 0.403 & 1.261 & 0.163 & 0.303 & 0.926 & 0.150 \\
No target in DiT, mask & 0.359 & 1.094 & 0.142 & 0.383 & 1.150 & 0.163 & 0.391 & 1.210 & 0.125 & 0.302 & 0.921 & 0.138 \\
No target in DiT, shuffled mask & 0.374 & 1.155 & 0.163 & 0.406 & 1.249 & 0.175 & 0.411 & 1.285 & 0.175 & 0.305 & 0.932 & 0.138 \\
All image tokens removed & 0.456 & 1.418 & 0.242 & 0.469 & 1.449 & 0.263 & 0.507 & 1.601 & 0.275 & 0.393 & 1.204 & 0.188 \\
Black input frames & 0.486 & 1.564 & 0.304 & 0.518 & 1.598 & 0.362 & 0.561 & 1.882 & 0.375 & 0.379 & 1.211 & 0.175 \\
\midrule
Baseline, black frames & 0.441 & 1.408 & 0.246 & 0.528 & 1.602 & 0.300 & 0.476 & 1.568 & 0.287 & 0.318 & 1.052 & 0.150 \\
\bottomrule
\end{tabular}
\end{table}

\begin{table}[!htb]
\centering\tiny
\caption{Turn response metrics at step 38{,}400. Direction accuracy and lateral displacement ratios are computed over left/right turn samples with $|y_\mathrm{GT}|>0.5$\,m.}
\label{tab:app_turn}
\begin{tabular}{lccccccccc}
\toprule
 & Baseline & Mask & No mask & Shuffled & Mirrored & No target & No img tokens & Black & Base.\ black \\
\midrule
Corr.\ of final lateral offset & 0.986 & 0.987 & 0.987 & 0.987 & 0.986 & 0.984 & 0.967 & 0.959 & 0.969 \\
Std of predicted final lateral (m) & 4.451 & 4.482 & 4.488 & 4.497 & 4.384 & 4.475 & 3.981 & 3.668 & 3.993 \\
Left: direction accuracy & 0.987 & 1.000 & 1.000 & 1.000 & 0.987 & 1.000 & 1.000 & 1.000 & 0.987 \\
Right: direction accuracy & 1.000 & 1.000 & 1.000 & 1.000 & 1.000 & 0.987 & 0.973 & 1.000 & 0.973 \\
Left: median pred./GT lateral & 0.993 & 1.002 & 1.000 & 0.993 & 0.961 & 0.984 & 0.919 & 0.897 & 0.931 \\
Right: median pred./GT lateral & 1.031 & 1.020 & 1.018 & 1.025 & 0.973 & 0.986 & 0.823 & 0.726 & 0.838 \\
\bottomrule
\end{tabular}
\end{table}

\begin{figure}[!htb]
\centering
\includegraphics[width=0.9\linewidth]{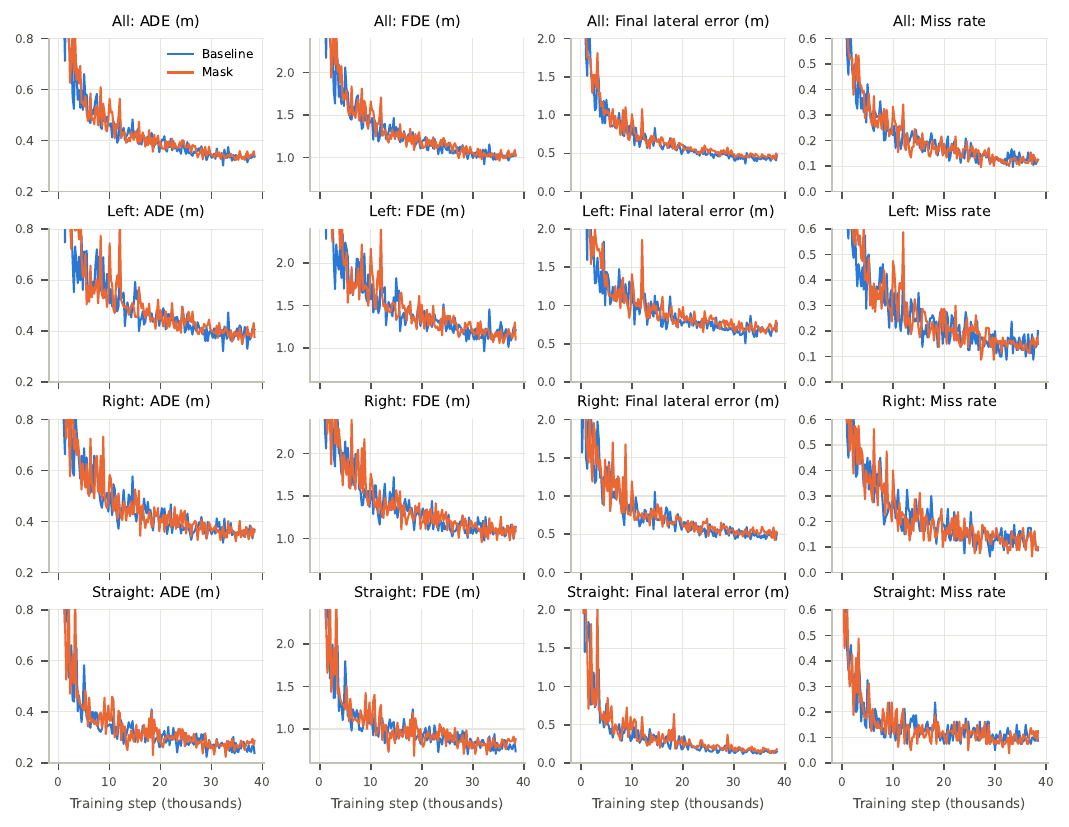}
\caption{Open-loop validation metrics evaluated every 250 training steps. Vertical axes are clipped for readability, placing initial early-training evaluations off-scale.}
\label{fig:curve_openloop}
\end{figure}

\begin{figure}[!htb]
\centering
\includegraphics[width=0.8\linewidth]{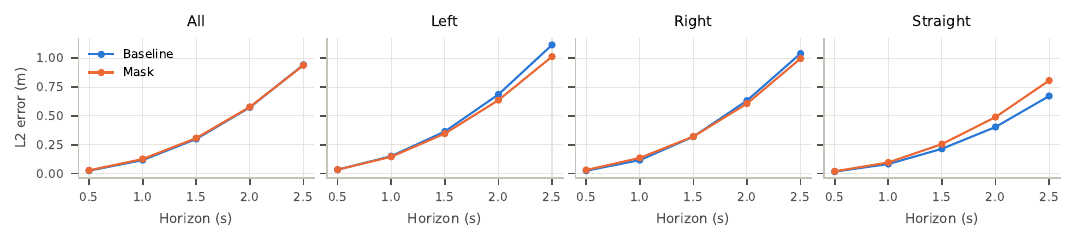}
\caption{L2 position error across the 2.5\,s prediction horizon at step 38{,}400.}
\label{fig:horizon}
\end{figure}

\begin{figure}[!htb]
\centering
\includegraphics[width=0.85\linewidth]{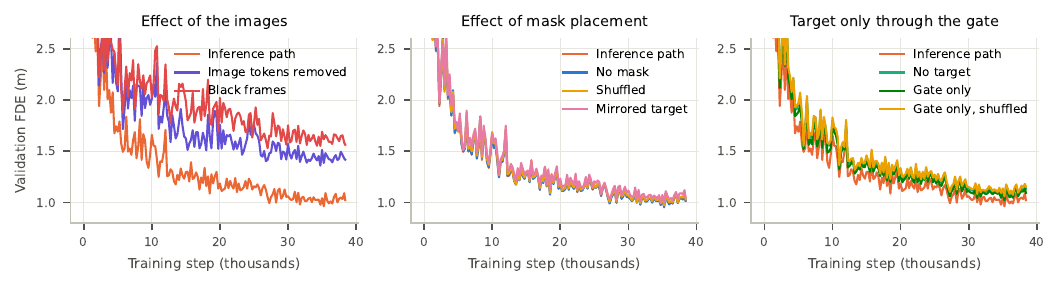}
\caption{Mask-run diagnostic controls across training (validation FDE). Control definitions are provided in Appendix~\ref{app:mask}.}
\label{fig:curve_controls}
\end{figure}

\begin{figure}[!htb]
\centering
\includegraphics[width=0.85\linewidth]{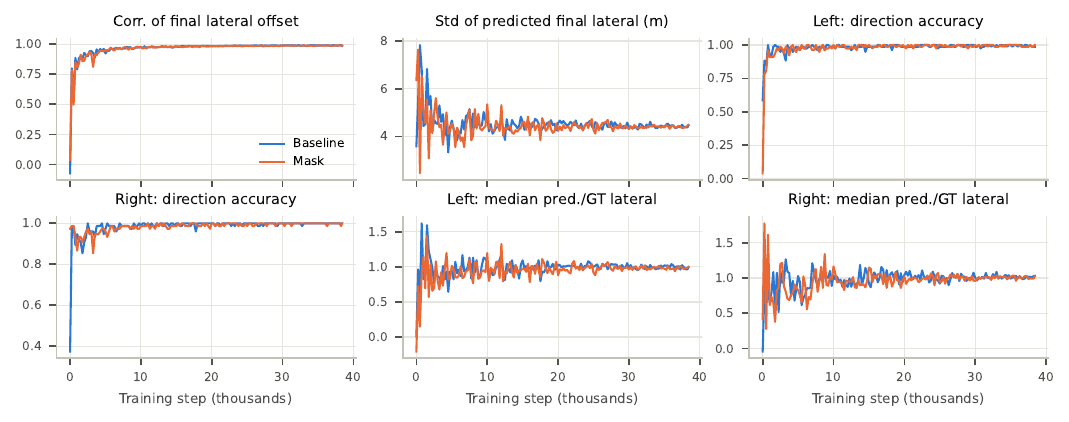}
\caption{Turn response metrics across training.}
\label{fig:curve_turn}
\end{figure}

\begin{figure}[!htb]
\centering
\includegraphics[width=0.85\linewidth]{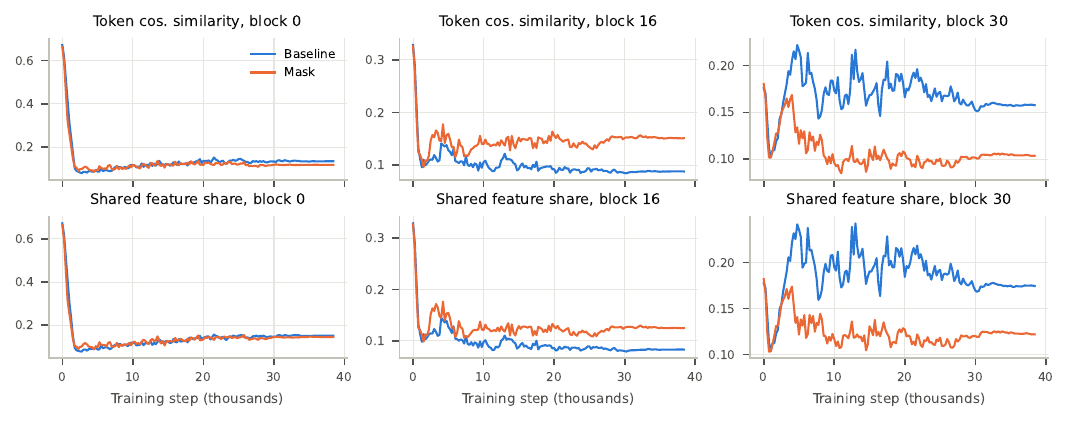}
\caption{Cosine similarity among projected VLM tokens entering the DiT at three depth blocks. In the deepest block, token representations in the masked run exhibit lower pairwise similarity than in the baseline.}
\label{fig:curve_features}
\end{figure}

\begin{figure}[!htb]
\centering
\includegraphics[width=0.85\linewidth]{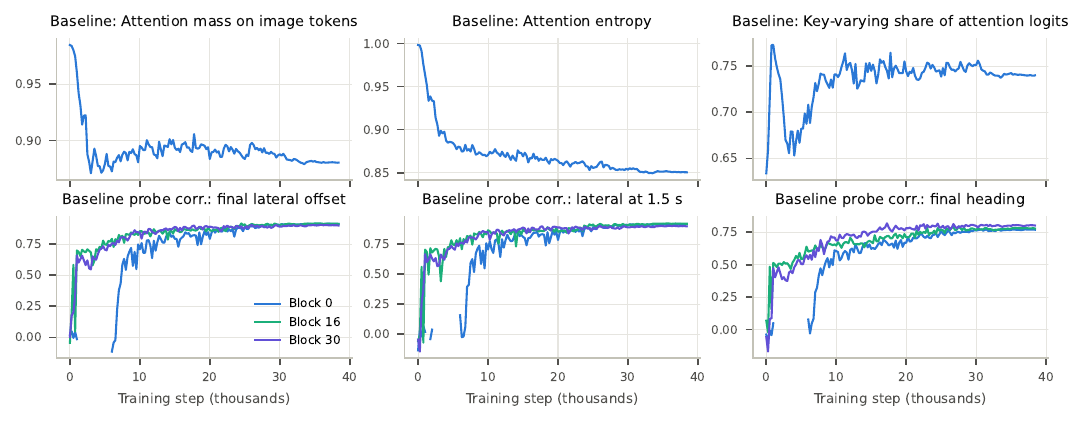}
\caption{Baseline diagnostics: DiT cross-attention statistics (top), and correlations between detached probes (attention pooling and an MLP) on projected VLM features and future motion across three DiT blocks (bottom).}
\label{fig:curve_baseline_diag}
\end{figure}

\begin{figure}[!htb]
\centering
\includegraphics[width=0.85\linewidth]{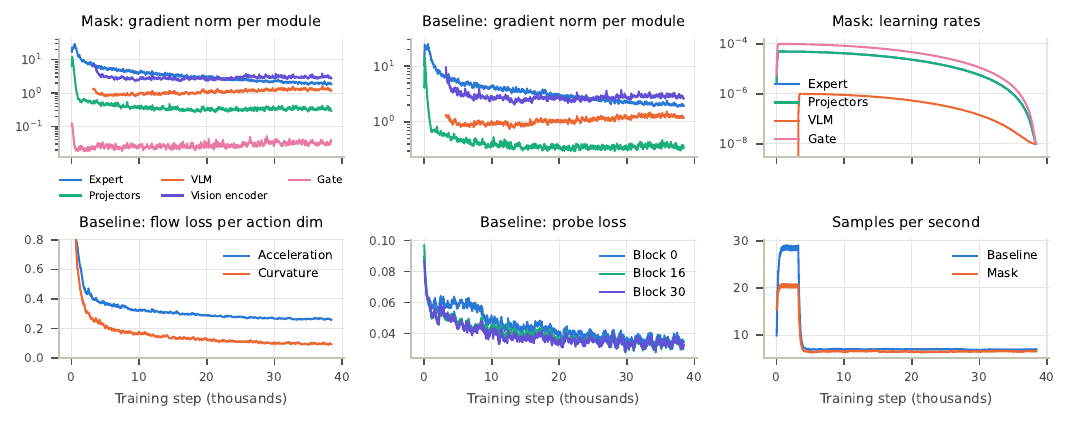}
\caption{Smoothed optimization signals across training. Throughput is substantially higher during the first epoch while the VLM backbone is frozen.}
\label{fig:curve_optim}
\end{figure}

\FloatBarrier
\section{Driver pipeline}
\label{app:driver}
\textbf{Observation handling.} The driver service implements AlpaSim's gRPC API. Incoming JPEG-compressed camera frames are decoded upon arrival, with only the front wide-angle camera feeding the policy and object detector. Camera calibration is derived from the session's nominal $f$-$\theta$ intrinsics (scaled to the decoded frame resolution), falling back to the static rig configuration file if unavailable. Egomotion states are recorded into the latest observation bundle at every \texttt{Drive} call, ensuring that the 0.5\,s replanning cycle always reads a complete 10\,Hz state history.

\textbf{Route.} The challenge evaluation presets use map-based routes with a 40\,m start offset. Whenever the route's initial waypoint lies beyond the resampling step, a cubic Hermite spline departing the rig origin along $+x$ and joining the route's first segment bridges the gap; the resulting polyline is then linearly extended beyond the farthest target lookahead distance and uniformly resampled at 0.1\,m intervals. The target point is extracted exactly as in training, substituting the resampled map route for the logged future trajectory.

\textbf{Serving.} Each generated plan is anchored in the local coordinate frame at creation time. Every control tick emits the planned waypoints corresponding to timestamps at or after the current time on its 10\,Hz grid. If a new plan cannot be computed (due to a missing camera frame, missing ego pose, or runtime planning exception), the previously cached plan remains active; if no valid cached plan exists, the driver falls back to serving a straight-line trajectory at constant speed $\max(2\,\mathrm{m/s},v)$.

\textbf{AEB at a tick.} At each control tick, the AEB module executes five steps: (1) If the newest front-camera frame has not yet been processed and a detection is scheduled, run the detector and project the bounding boxes onto the ground plane. (2) Transform the remainder of the cached plan into the current rig frame and identify the nearest lead obstacle along it. (3) Update the obstacle tracker and compute the longitudinal speed cap. (4) When a speed cap is active, let $w(\tau)$ denote the distance that the cached plan still intends to cover over $\tau$ seconds, and let $u(\tau)$ denote the maximum distance permitted by the deceleration ramp (or $v\tau$ under the no-acceleration condition). The served trajectory starts at the ego vehicle's orthogonal projection onto the planned path, advancing to arc-length positions $s_\mathrm{ego}+\min(w,u)$ on a 0.1\,s grid with yaw angles interpolated along the geometric path. (5) If the speed cap has just been released on the current tick, hold the current speed for one tick and invalidate the cached plan so that the policy replans on the next tick. Without Step~(5), reverting directly to the cached plan---whose timestamped waypoints now lie far ahead of the braked ego vehicle---would cause the tracking controller to accelerate aggressively toward the obstacle it had just braked for.

\FloatBarrier
\section{AEB development}
\label{app:aeb}
\textbf{Why it was added.} Without AEB, the single largest source of score loss is at-fault collisions---predominantly front-end collisions with lead vehicles traveling substantially slower than the ego vehicle. Across 178 first-incident front collisions pooled from five evaluation runs (mask, TE4, hist\_ft, and two sig02 rollouts), the median ego impact speed is 11.9\,m/s (interquartile range: 6.8--17.8\,m/s). Kinematic profiles over the 3\,s preceding impact differ only moderately from non-incident 3\,s windows prior to any incident (6{,}077 windows): a speed increase exceeding 2\,m/s occurs in 35\% vs.\ 25\% of windows; mean acceleration exceeding 0.3\,m/s$^2$ occurs in 48\% vs.\ 37\%; and peak acceleration exceeding 1.5\,m/s$^2$ occurs in 24\% vs.\ 21\%. Imposing a global acceleration cap would therefore mostly cost route progress, motivating an obstacle-aware braking rule. Because the driver receives no privileged state for other traffic agents and ground-truth perception labels are disabled in the challenge presets, obstacles must be detected directly from camera images.

\textbf{Offline replay.} Each evaluation rollout log stores the front-camera JPEGs, ego states, and served trajectory plans. We executed the detector across these logged frames inside the driver container, cached the raw bounding boxes, and tuned the AEB decision rules via offline replay. An initial configuration (no bounding-box bottom-edge shrinkage, 1.6\,m corridor half-width, 2\,m/s$^2$ trigger threshold) triggered braking at least 1\,s prior to impact in 38 of 41 front collisions, but also triggered false-positive braking in 29 of 46 sampled clean rollouts. Visual inspection traced these false positives primarily to cars parked along the curb: when viewed obliquely, the bottom corners of a vehicle's 2D bounding box extend laterally beyond the physical wheels and intrude into the planned driving corridor. Table~\ref{tab:app_sweep} illustrates the resulting trade-off across hyperparameter settings. Note that this offline replay is open-loop and tuned on the local evaluation scenes.

\begin{table}[!htb]
\centering\scriptsize
\caption{Offline AEB hyperparameter sweep on replayed detections from the masked policy's 150 rollouts (comprising 41 front collisions and 62 incident-free rollouts). A braking event is counted only when the speed cap shortens the planned path at a 1\,s horizon by more than 0.3\,m. The selected deployment configuration is highlighted in bold. Evaluation is open-loop (ego motion does not react to simulated braking).}
\label{tab:app_sweep}
\begin{tabular}{cccccccc}
\toprule
Shrink & Half-width & Trigger & Gap & Samples & Hits, $\ge1$\,s & Hits, $\ge0.5$\,s & Clean: rollouts / frames \\
\midrule
0.0 & 1.6 & 3.0 & 2.5 & 4 & 36 & 38 & 38 / 769 \\
0.3 & 1.3 & 4.0 & 1.5 & 6 & 24 & 28 & 15 / 92 \\
\textbf{0.3} & \textbf{1.3} & \textbf{3.0} & \textbf{2.5} & \textbf{6} & \textbf{29} & \textbf{32} & \textbf{19} / \textbf{164} \\
0.3 & 1.3 & 4.0 & 2.5 & 6 & 29 & 32 & 17 / 127 \\
0.15 & 1.3 & 3.0 & 2.5 & 6 & 32 & 37 & 22 / 227 \\
0.15 & 1.45 & 3.0 & 2.5 & 6 & 34 & 38 & 24 / 268 \\
0.3 & 1.6 & 3.0 & 2.5 & 6 & 31 & 36 & 22 / 214 \\
0.0 & 1.3 & 4.0 & 1.5 & 6 & 28 & 34 & 24 / 234 \\
0.3 & 1.3 & 2.0 & 2.5 & 6 & 31 & 32 & 23 / 239 \\
\bottomrule
\end{tabular}
\end{table}

\textbf{Chosen configuration.} Under the selected parameter setting, the 3\,s time-to-collision no-acceleration rule introduces 47 additional frames of intervention across clean rollouts. In the AlpaSim scoring metric, being rear-ended by a trailing agent does not zero the scene score (as only front and lateral impacts are penalized as at-fault) and route progress saturates at 80\% of the recorded reference distance; consequently, the challenge metric rewards this conservative braking trade-off. However, as noted in Section~\ref{sec:limits}, inducing rear-end collisions would pose a severe safety hazard in real-world driving.

\textbf{Detector choice.} RT-DETRv2 is natively supported in our container image's \texttt{transformers} library with Apache-2.0 licensed weights, and its DETR-style set prediction architecture requires no post-processing non-maximum suppression (NMS). Profiling the R18 and R50 backbones inside the container on a heavily loaded NVIDIA RTX PRO 6000 Blackwell GPU yielded median per-frame latencies of approximately 38\,ms and 49\,ms, with peak VRAM allocations of 183\,MiB and 298\,MiB, respectively. Offloading image resizing to the CPU and enabling \texttt{fp16} autocast did not substantially alter R18 latency (approximately 35\,ms), and processing a batch of 4 frames required nearly the same wall-clock time as a batch of 1 (approximately 30\,ms), indicating that CPU-side kernel launch overhead dominated under multi-tenant host contention. Attempting to capture the detector within a CUDA graph failed because the upstream implementation performs CPU-to-GPU tensor writes during the \texttt{forward} pass.

\FloatBarrier
\section{Closed-loop results in full}
\label{app:closed}
\textbf{Protocol.} Evaluations were conducted on a fixed 150-scene subset of the 441-scene \texttt{nurec\_curated\_val} benchmark using the \texttt{+e2e\_challenge=dev} preset, configured with a multi-GPU topology and driver video recording enabled on shared multi-tenant hosts. The first rollout sets (baseline, mask, and 10\,Hz AEB), as well as the TE4, hist\_ft, and 8-sample variants, were executed on an NVIDIA RTX PRO 6000 Blackwell host (8 GPUs). The second rollout sets, along with the sig02 and 5\,Hz AEB variants, were executed on NVIDIA RTX 6000 Ada hosts (3 GPUs each). Because the first and second rollouts for each configuration ran on distinct hardware architectures, discrepancies between the two rollouts reflect both hardware differences and stochastic run-to-run variance. Any scenes that terminated prematurely due to renderer out-of-memory errors on shared GPUs were re-run with reduced concurrency per renderer instance. Scoring terminates at the first collision or off-road departure: a rollout receives a score of 0 if it incurs an at-fault collision (front or lateral impact), an off-road violation, or a lateral corridor exit (deviating by more than 4\,m from the recorded reference path), and $\min(\mathrm{progress}/0.8,1)$ otherwise.

\begin{figure}[!htb]
\centering
\includegraphics[width=0.8\linewidth]{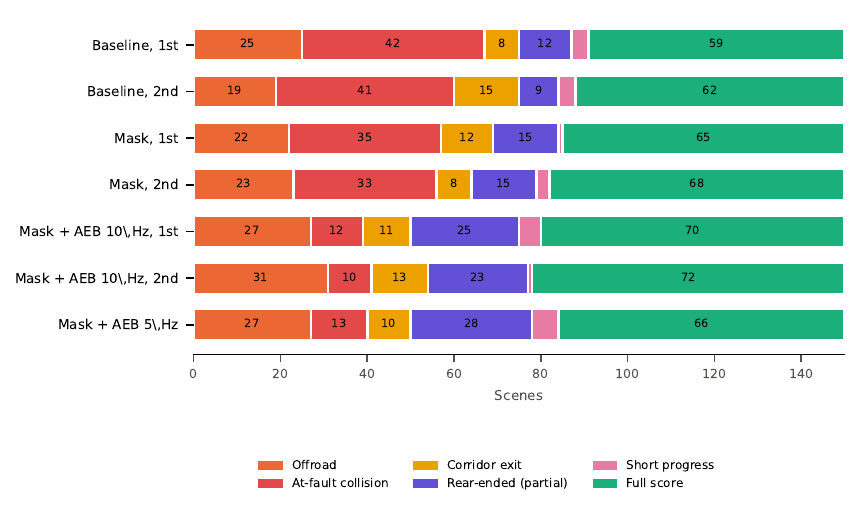}
\caption{Per-scene rollout outcomes across the same 150 evaluation scenes.}
\label{fig:outcomes}
\end{figure}

\begin{table}[!htb]
\centering\tiny
\caption{Score-loss decomposition across the same 150 scenes in every column, formatted as: affected scenes / contribution to $(1-\text{mean score})$.}
\label{tab:app_loss}
\begin{tabular}{lccccccc}
\toprule
 & Baseline, 1st & Baseline, 2nd & Mask, 1st & Mask, 2nd & AEB 10\,Hz, 1st & AEB 10\,Hz, 2nd & AEB 5\,Hz \\
\midrule
Offroad & 25 / 0.167 & 19 / 0.127 & 22 / 0.147 & 23 / 0.153 & 27 / 0.180 & 31 / 0.207 & 27 / 0.180 \\
At-fault collision & 42 / 0.280 & 41 / 0.273 & 35 / 0.233 & 33 / 0.220 & 12 / 0.080 & 10 / 0.067 & 13 / 0.087 \\
Corridor exit & 8 / 0.053 & 15 / 0.100 & 12 / 0.080 & 8 / 0.053 & 11 / 0.073 & 13 / 0.087 & 10 / 0.067 \\
Rear-ended (partial) & 12 / 0.022 & 9 / 0.015 & 15 / 0.038 & 15 / 0.037 & 25 / 0.054 & 23 / 0.046 & 28 / 0.067 \\
Short progress & 4 / 0.002 & 4 / 0.003 & 1 / 0.002 & 3 / 0.002 & 5 / 0.005 & 1 / 0.002 & 6 / 0.009 \\
Full score & 59 / 0.000 & 62 / 0.000 & 65 / 0.000 & 68 / 0.000 & 70 / 0.000 & 72 / 0.000 & 66 / 0.000 \\
\midrule
Mean score & 0.476 & 0.481 & 0.499 & 0.534 & 0.607 & 0.591 & 0.590 \\
Mean progress & 0.745 & 0.773 & 0.737 & 0.739 & 0.737 & 0.732 & 0.720 \\
Mean Drive (s) & 0.133 & 0.066 & 0.110 & 0.146 & 0.201 & 0.085 & 0.264 \\
Host & Blackwell & Ada & Blackwell & Ada & Blackwell & Ada & Ada \\
\bottomrule
\end{tabular}
\end{table}

\begin{table}[!htb]
\centering\scriptsize
\caption{Paired per-scene comparisons across the same 150 scenes. CI denotes the 95\% bootstrap confidence interval (10{,}000 resamples); $p$ denotes the two-sided sign test $p$-value (ignoring ties, uncorrected for multiple comparisons); (2) indicates the per-scene mean across two rollouts.}
\label{tab:app_pairs}
\begin{tabular}{lccccccc}
\toprule
Comparison & $n$ & Ref. & Run & Diff. & 95\% CI & W/L & $p$ \\
\midrule
Mask (2) vs.\ baseline (2) & 150 & 0.479 & 0.517 & +0.038 & [-0.018, +0.093] & 37/26 & 0.21 \\
Mask 1st vs.\ baseline 1st & 150 & 0.476 & 0.499 & +0.023 & [-0.049, +0.093] & 28/20 & 0.31 \\
Mask 2nd vs.\ baseline 2nd & 150 & 0.481 & 0.534 & +0.053 & [-0.022, +0.126] & 31/19 & 0.12 \\
Mask 2nd vs.\ mask 1st & 150 & 0.499 & 0.534 & +0.035 & [-0.017, +0.089] & 22/16 & 0.42 \\
Baseline 2nd vs.\ baseline 1st & 150 & 0.476 & 0.481 & +0.005 & [-0.065, +0.076] & 25/20 & 0.55 \\
AEB 10\,Hz (2) vs.\ mask (2) & 150 & 0.517 & 0.599 & +0.082 & [+0.023, +0.143] & 41/28 & 0.15 \\
AEB 10\,Hz 1st vs.\ mask 1st & 150 & 0.499 & 0.607 & +0.108 & [+0.038, +0.176] & 37/14 & $<$0.01 \\
AEB 10\,Hz 2nd vs.\ mask 2nd & 150 & 0.534 & 0.591 & +0.057 & [-0.024, +0.137] & 34/26 & 0.37 \\
AEB 10\,Hz 2nd vs.\ 1st & 150 & 0.607 & 0.591 & -0.016 & [-0.087, +0.054] & 24/30 & 0.50 \\
AEB 5\,Hz vs.\ mask (2) & 150 & 0.517 & 0.590 & +0.073 & [+0.012, +0.134] & 43/22 & 0.01 \\
AEB 5\,Hz vs.\ mask 2nd & 150 & 0.534 & 0.590 & +0.056 & [-0.013, +0.125] & 36/21 & 0.06 \\
AEB 5\,Hz vs.\ AEB 10\,Hz (2) & 150 & 0.599 & 0.590 & -0.009 & [-0.065, +0.044] & 34/34 & 1.00 \\
\bottomrule
\end{tabular}
\end{table}

\FloatBarrier
\begin{wraptable}{r}{0.6\linewidth}
\centering\scriptsize
\caption{Scene outcome transition matrix when adding 10\,Hz AEB (first rollouts, same host, 150 scenes): rows denote outcomes without AEB, and columns denote outcomes with AEB.}
\label{tab:app_flow}
\begin{tabular}{@{}lcccccc@{}}
\toprule
Without $\backslash$ with & Offroad & Coll. & Exit & Rear & Short & Full \\
\midrule
Offroad & 19 &  & 1 &  &  & 2 \\
At-fault collision & 2 & 9 & 2 & 11 & 2 & 9 \\
Corridor exit & 3 &  & 5 & 1 &  & 3 \\
Rear-ended (partial) & 1 &  & 1 & 11 & 2 &  \\
Short progress &  &  &  &  & 1 &  \\
Full score & 2 & 3 & 2 & 2 &  & 56 \\
\bottomrule
\end{tabular}
\end{wraptable}
\textbf{Where the collisions go.} Table~\ref{tab:app_flow} follows each scene through the first rollouts on the same host. Of the 35 scenes with an at-fault collision without AEB, 9 remain at-fault collisions, 9 reach the full-score outcome, and 11 become rear-ended scenes, while 4 turn into lateral failures (2 off-road, 2 corridor exits) and 2 end with short progress. In the other direction, of the 65 full-score scenes without AEB, 56 stay at full score, 3 gain an at-fault collision, 4 gain a lateral failure, and 2 are rear-ended. Scenes that were already off-road without AEB mostly stay so (19 of 22).

\textbf{Collision kinematics without AEB.} Table~\ref{tab:app_kin} compares pre-collision kinematic statistics across model variants. Among the first-incident front collisions without AEB, the ego vehicle accelerated by more than 2\,m/s over the preceding 3\,s in \RMaskAccel\% (mask) vs.\ \RBaseAccel\% (baseline) of cases, and the final planned trajectory failed to command deceleration in \RMaskNoSlow\% vs.\ \RBaseNoSlow\% of cases.

\begin{table}[!htb]
\centering\scriptsize
\caption{Pre-impact kinematics for first-incident front collisions across the same 150 scenes: overall front-collision rate; fraction with a speed increase $>2$\,m/s over the preceding 3\,s; fraction whose final planned trajectory failed to command deceleration (planned speed 2\,s ahead $\ge v-0.3$\,m/s); fraction where the nearest lead actor was traveling below 1\,m/s (approximate); and median ego impact speed (m/s).}
\label{tab:app_kin}
\begin{tabular}{lcccccc}
\toprule
 & Hits & Rate & Accel. & No slowing & Stopped lead & $v_\mathrm{hit}$ \\
\midrule
Baseline (2) & 90/301 & 29.9\% & 42\% & 68\% & 39\% & 11.6 \\
Mask (2) & 77/300 & 25.7\% & 30\% & 64\% & 38\% & 10.9 \\
Mask + AEB 10\,Hz (2) & 28/300 & 9.3\% & 32\% & 50\% & 61\% & 9.1 \\
Mask + AEB 5\,Hz & 12/150 & 8.0\% & 17\% & 42\% & 67\% & 2.4 \\
hist\_ft & 46/147 & 31.3\% & 46\% & 67\% & 39\% & 13.2 \\
sig02 & 50/150 & 33.3\% & 32\% & 62\% & 36\% & 12.9 \\
Mask + TE4 & 41/150 & 27.3\% & 29\% & 66\% & 34\% & 10.5 \\
Baseline + TE4 & 34/148 & 23.0\% & 35\% & 62\% & 44\% & 9.9 \\
\bottomrule
\end{tabular}
\end{table}

\begin{figure}[!htb]
\centering
\includegraphics[width=\linewidth]{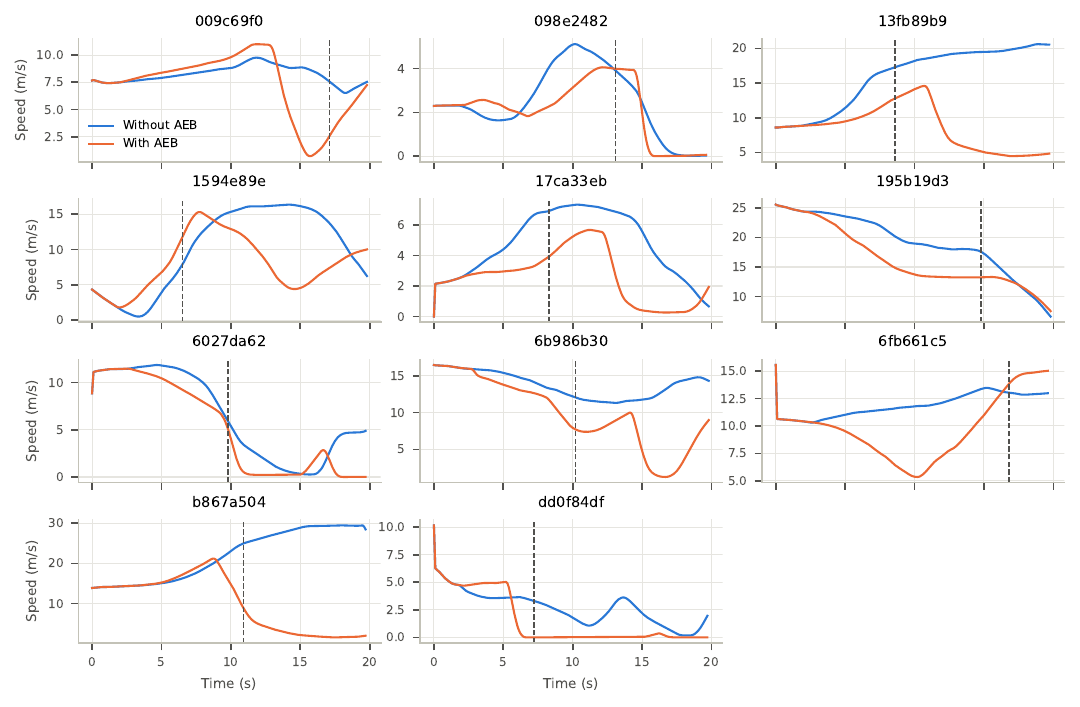}
\caption{Ego speed profiles across the 11 scenes that incurred an at-fault collision without AEB but achieved a full score with 10\,Hz AEB enabled (first rollouts). Dashed vertical lines mark the time of collision without AEB.}
\label{fig:aeb_speed}
\end{figure}

\begin{figure}[!htb]
\centering
\includegraphics[width=\linewidth]{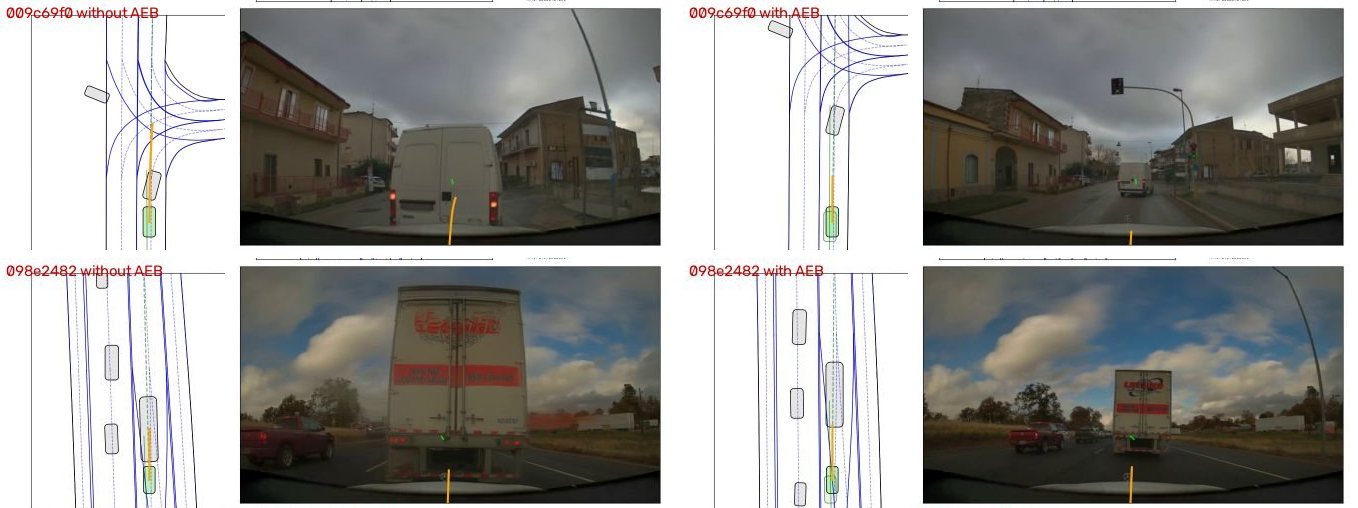}
\caption{Side-by-side comparison 0.3\,s prior to collision without AEB versus the same elapsed rollout time with AEB enabled, for 2 of the 11 rescued scenes. Each tile shows the map view (left) and the camera view (right).}
\label{fig:aeb_pairs}
\end{figure}

\textbf{Remaining collisions with AEB (first rollout).} Table~\ref{tab:app_hits} details all 12 remaining at-fault collisions under 10\,Hz AEB. \emph{Corridor}: the obstacle remained outside the 1.3\,m lateral detection corridor until shortly before impact (or throughout the approach); relaxing the geometry to zero edge shrinkage and a 2\,m half-width in offline replay detects most of these obstacles earlier, but at the cost of the false-positive braking events analyzed in Table~\ref{tab:app_sweep}. \emph{Side/behind}: the colliding agent approached from outside the forward corridor (e.g., a crossing bus or a vehicle overtaking from the rear-left). \emph{Close range}: the ground-projected distance estimate stagnated near 4.4\,m, preventing the linear regression tracker from observing a positive closing speed; we hypothesize (though have not formally verified) that the ego vehicle's hood occluded the lead vehicle's wheels at close range. \emph{Late brake}: in scene \texttt{1316f5fa}, a step discontinuity in estimated distance reset the obstacle track, delaying braking until approximately 0.5\,s before impact; in scene \texttt{382a7247}, AEB decelerated the ego vehicle from 9.2\,m/s down to 0.7\,m/s, but low-speed contact still occurred.

\begin{table}[!htb]
\centering\tiny
\caption{Breakdown of the 12 remaining at-fault collisions under 10\,Hz AEB (first rollout). Columns report: elapsed time $t$ from rollout start (s); ego speeds 3\,s prior ($v_{-3}$) and at impact ($v$) in m/s; minimum acceleration $a_\mathrm{min}$ over the last 3\,s (m/s$^2$); kinematics of the nearest traffic agent at impact (speed $v_o$, longitudinal and lateral center offsets in the ego frame, and relative heading in degrees; approximate, as the nearest agent may not be the struck actor); and lead time (s) prior to impact at which the AEB trigger condition was first met in offline detector replay. Failure categories reflect manual inspection.}
\label{tab:app_hits}
\begin{tabular}{llcccccccccc}
\toprule
Scene & Class & Type & $t$ & $v_{-3}$ & $v$ & $a_\mathrm{min}$ & $v_o$ & Lon. & Lat. & Head. & Brake \\
\midrule
\texttt{0caa8f1a} & Close range & front & 9.4 & 0.1 & 1.2 & -1.5 & 0.5 & 4.4 & -1.4 & -10 & 4.5 \\
\texttt{d641893f} & Close range & front & 15.1 & 0.9 & 4.2 & 0.8 & 12.0 & -1.1 & 3.1 & -179 & none \\
\texttt{05855d57} & Corridor & front & 11.5 & 30.3 & 32.0 & -0.7 & 26.8 & 4.3 & -0.0 & 2 & 0.3 \\
\texttt{213dfdac} & Corridor & front & 3.2 & 11.3 & 8.7 & -1.5 & 0.1 & 2.2 & -1.9 & 7 & none \\
\texttt{24a50fcc} & Corridor & front & 11.3 & 9.8 & 9.7 & -0.7 & 7.1 & 1.9 & -1.9 & 4 & none \\
\texttt{8b25a6ac} & Corridor & side & 10.6 & 9.9 & 9.9 & -0.1 & 0.0 & 2.7 & -2.1 & -1 & none \\
\texttt{d5f0f729} & Corridor & front & 18.2 & 24.3 & 21.7 & -1.5 & 0.1 & 4.2 & -1.6 & 1 & none \\
\texttt{fadc73da} & Corridor & front & 17.1 & 1.4 & 6.5 & -0.2 & 0.0 & 3.7 & 2.1 & 4 & none \\
\texttt{1316f5fa} & Late brake & front & 19.8 & 6.6 & 7.0 & -8.3 & 0.1 & 4.3 & -1.7 & -5 & 0.5 \\
\texttt{382a7247} & Late brake & front & 9.0 & 9.2 & 0.7 & -8.6 & 0.0 & 4.0 & 1.8 & -162 & 2.9 \\
\texttt{0d79a6a0} & Side/behind & side & 3.2 & 10.5 & 8.1 & -5.5 & 11.8 & 0.1 & 2.5 & -22 & 0.9 \\
\texttt{1d879259} & Side/behind & front & 6.0 & 2.1 & 4.6 & -2.9 & 6.0 & 4.6 & -0.8 & -71 & 0.3 \\
\bottomrule
\end{tabular}
\end{table}

\begin{figure}[!htb]
\centering
\includegraphics[width=\linewidth]{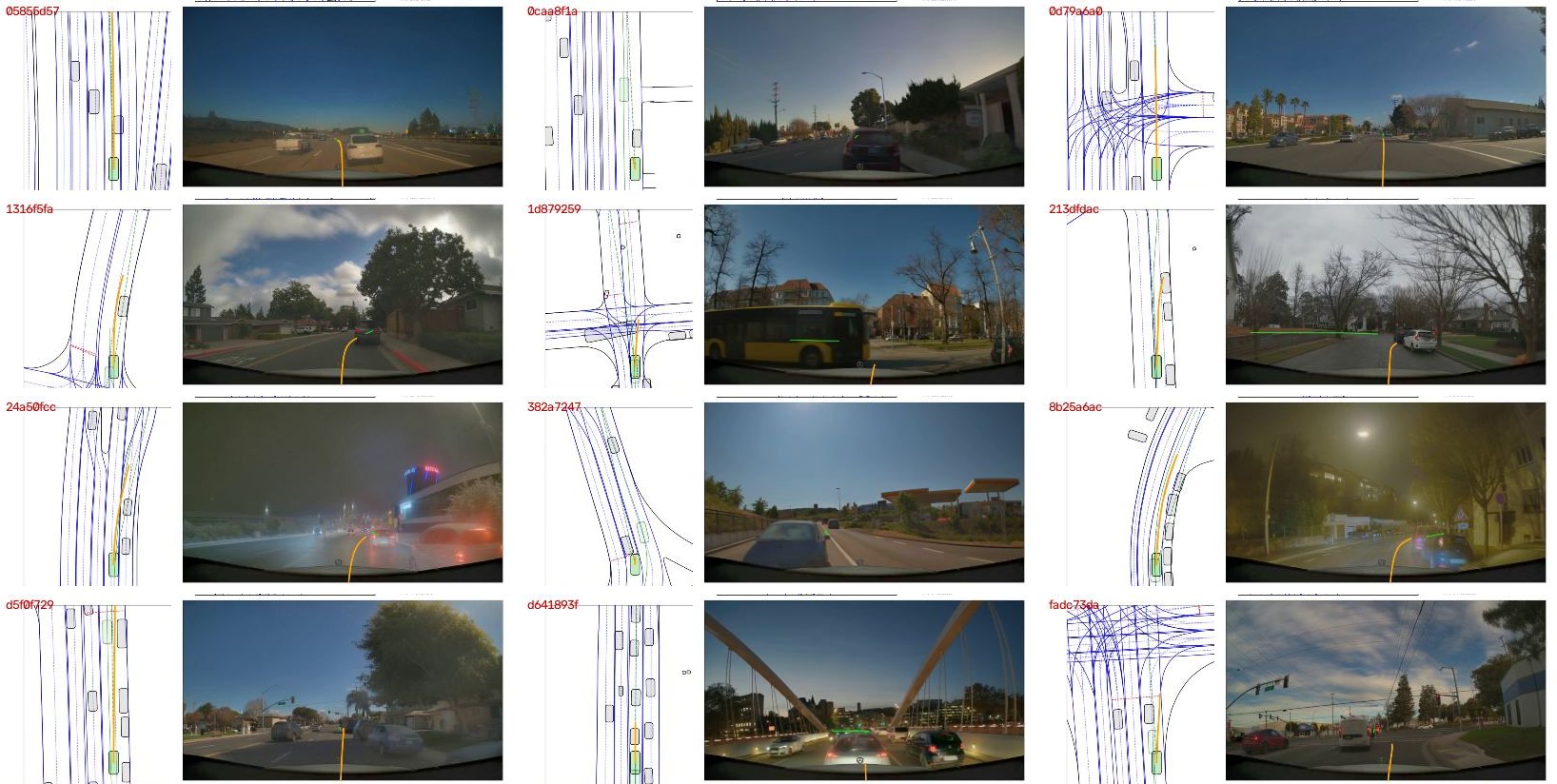}
\caption{Views 1\,s prior to impact for all 12 remaining at-fault collisions under 10\,Hz AEB (first rollout); Map view (left) and camera view (right); Table~\ref{tab:app_hits} classifies them.}
\label{fig:sheet_collisions}
\end{figure}

\textbf{Lateral failures with AEB (first rollout).} Among the 27 scenes resulting in off-road violations under AEB, 19 also went off-road in the first rollout without AEB, and only 3 experienced hard braking (deceleration below $-2$\,m/s$^2$) during the preceding 3\,s, indicating that AEB intervention is not the primary cause of off-road departures. Eighteen of the 27 scenes exhibited a cumulative heading change of less than $15^\circ$ over the final 3\,s, consistent with gradual lateral drift, and video inspection of 12 sampled scenes frequently shows the planned trajectory drifting toward the right shoulder. Among the 31 lateral failures (off-road or corridor exit) with a well-defined lateral offset direction, 17 occurred to the right of the route and 9 to the left, while 5 occurred within 0.3\,m of the route centerline.

\begin{wraptable}{r}{0.48\linewidth}
\centering\scriptsize
\caption{Offline evaluation of a hypothetical route-distance guard (not deployed): number of lateral failure rollouts flagged at least 1\,s prior to failure versus number of full-score clean rollouts falsely flagged at any point.}
\label{tab:app_guard}
\begin{tabular}{@{}ccccc@{}}
\toprule
 & \multicolumn{2}{c}{Ego} & \multicolumn{2}{c}{Plan at 2\,s} \\
Thr.\ (m) & Fail. & Clean & Fail. & Clean \\
\midrule
0.5 & 27/38 & 50/70 & 29/38 & 49/70 \\
0.75 & 21/38 & 37/70 & 26/38 & 39/70 \\
1.0 & 19/38 & 30/70 & 23/38 & 33/70 \\
1.5 & 15/38 & 17/70 & 17/38 & 20/70 \\
2.0 & 13/38 & 15/70 & 15/38 & 16/70 \\
3.0 & 8/38 & 7/70 & 12/38 & 9/70 \\
\bottomrule
\end{tabular}
\end{wraptable}
\textbf{Route as a lateral reference.} Because provided routes are generated from map lane centerlines, they represent a natural candidate reference for lateral safety guarding. When sampling route waypoints every 2\,m and placing the ego vehicle footprint along the route, only 114 of 36{,}843 sampled poses (0.3\%) trigger an off-road violation under the simulator's scoring logic (5{,}728 poses exceed a single lane's boundaries without intersecting a road edge). Thus, while the map route lies almost entirely within drivable road boundaries, Tables~\ref{tab:app_guard} and~\ref{tab:app_route} demonstrate that lateral distance to the route poorly discriminates impending lateral failures from normal driving variations; consequently, we did not deploy a route-distance guard.

\begin{table}[!htb]
\centering\scriptsize
\caption{Lateral distance (m) from the ego vehicle to the served map route prior to lateral failures under 10\,Hz AEB. Scenes lacking a covering route segment at the queried timestamp are omitted. For reference, across all frames of the 70 full-score rollouts, the lateral distance distribution has a median of 0.29\,m, 90th percentile of 1.14\,m, and 99th percentile of 2.99\,m.}
\label{tab:app_route}
\begin{tabular}{llcccccc}
\toprule
Failure & Before & $n$ & Median & p25 & p75 & $>1$\,m & $>2$\,m \\
\midrule
Offroad & $-3$\,s & 18 & 0.40 & 0.18 & 2.56 & 5 & 5 \\
 & $-2$\,s & 23 & 0.49 & 0.31 & 1.36 & 6 & 4 \\
 & $-1$\,s & 23 & 0.70 & 0.33 & 1.91 & 11 & 5 \\
 & $-0$\,s & 23 & 0.67 & 0.40 & 1.99 & 8 & 6 \\
\midrule
Corridor exit & $-3$\,s & 5 & 0.81 & 0.70 & 0.87 & 1 & 1 \\
 & $-2$\,s & 7 & 0.79 & 0.34 & 2.77 & 3 & 3 \\
 & $-1$\,s & 8 & 0.82 & 0.26 & 2.36 & 4 & 3 \\
 & $-0$\,s & 8 & 2.80 & 0.61 & 3.43 & 5 & 5 \\
\bottomrule
\end{tabular}
\end{table}

\begin{figure}[!htb]
\centering
\includegraphics[width=\linewidth]{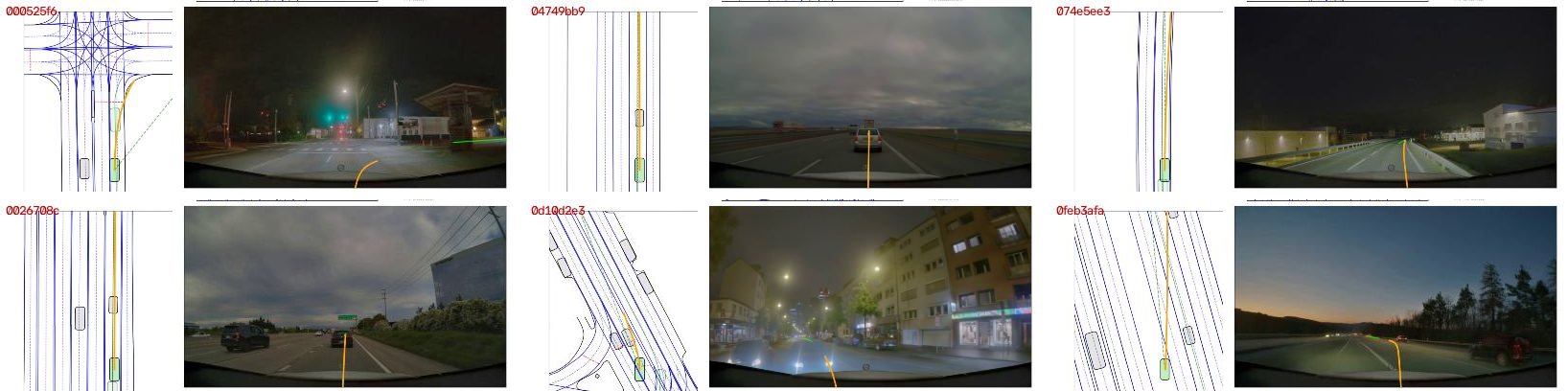}
\caption{Views 0.8\,s prior to the event under 10\,Hz AEB (first rollout), in scene-ID order: the first 3 of the 27 off-road failures (top) and the first 3 of the 11 lateral corridor exits (bottom). Map view (left) and camera view (right) of each scene.}
\label{fig:sheet_lateral}
\end{figure}

\FloatBarrier
\section{Variants evaluated and not submitted}
\label{app:others}
All additional variants were evaluated using the same local closed-loop protocol without AEB. We report these experiments as negative or inconclusive results:
\begin{itemize}\setlength\itemsep{0pt}
\item \textbf{Temporal ensembling (TE4).} Following the temporal ensembling scheme of ACT~\cite{act}, each newly predicted trajectory plan is averaged with up to 3 preceding plans across their overlapping time horizons; each older plan is rigidly transformed so that its pose at the new plan's creation timestamp aligns with the current ego pose, using exponential decay weights $\exp(-0.5\cdot\mathrm{age})$.
\item \textbf{8 samples.} Averaging 8 independent flow-matching trajectory samples per planning call within the normalized action space.
\item \textbf{hist\_ft.} A 3-epoch masked fine-tuning run initialized from the baseline's final checkpoint, incorporating ego-history dropout ($p=0.5$) with a binary presence flag (zeroing historical $x, y$, heading, speed, and all $a_\mathrm{lon}$ inputs while retaining current speed), past-motion dropout inspired by ChauffeurNet~\cite{chauffeurnet}, removal of fallback clip-endpoint targets, and an initial 300-step gate-only warmup phase.
\item \textbf{sig02.} Identical to \texttt{hist\_ft}, but replacing the bisection-budgeted mask with an unconstrained sigmoid mask (initialized to a mean retention rate of 0.2) regularized by an explicit squared budget penalty $(\mathrm{mean}_p\,m_p-\rho)^2$ of weight 0.1.
\end{itemize}
Both fine-tuning variants (\texttt{hist\_ft} and \texttt{sig02}) were trained for only 3 epochs.

\begin{table}[!htb]
\centering\scriptsize
\caption{Closed-loop evaluation of unsubmitted variants (without AEB), paired as in Table~\ref{tab:app_pairs} (comparing a single rollout of each variant against the two-rollout mean of the reference model).}
\label{tab:app_others}
\begin{tabular}{lccccccc}
\toprule
Comparison & $n$ & Ref. & Run & Diff. & 95\% CI & W/L & $p$ \\
\midrule
Baseline + TE4 vs.\ baseline (2) & 150 & 0.479 & 0.504 & +0.026 & [-0.027, +0.079] & 33/22 & 0.18 \\
Mask + TE4 vs.\ mask (2) & 150 & 0.517 & 0.495 & -0.021 & [-0.082, +0.041] & 25/29 & 0.68 \\
Mask + 8 samples vs.\ mask (2) & 150 & 0.517 & 0.512 & -0.005 & [-0.051, +0.040] & 24/21 & 0.77 \\
hist\_ft vs.\ mask (2) & 150 & 0.517 & 0.495 & -0.021 & [-0.082, +0.037] & 29/26 & 0.79 \\
sig02, 1st vs.\ mask (2) & 150 & 0.517 & 0.505 & -0.012 & [-0.077, +0.055] & 27/28 & 1.00 \\
sig02, 2nd vs.\ mask (2) & 150 & 0.517 & 0.497 & -0.020 & [-0.084, +0.043] & 29/29 & 1.00 \\
\bottomrule
\end{tabular}
\end{table}

\FloatBarrier
\section{Runtime measurements}
\label{app:runtime}
All local timing benchmarks were conducted on multi-tenant hosts shared with concurrent workloads and therefore differ from the official evaluation environment, which allocates 16 driver replicas across 4 GPUs with 2 concurrent rollouts per replica. Official runtime metrics are reported in the challenge submission status. Table~\ref{tab:app_loss} reports the end-to-end driver telemetry recorded during our local closed-loop evaluation runs (with video recording enabled and up to 12 concurrent rollouts per driver instance).

\begin{table}[!htb]
\centering\scriptsize
\caption{Standalone runtime of the pre-configured submission container images launched without arguments in the challenge sandbox: 2 replicas with 1 active session each on a single shared GPU over an approximately 27\,s rollout. All latencies are reported in seconds.}
\label{tab:app_sub}
\begin{tabular}{lccccc}
\toprule
 & Mean Drive & Planning call & Other calls, mean & Other calls, p50 & Calls with detector \\
\midrule
AEB 10\,Hz & 0.111 & 0.249 & 0.077 & 0.026 & 143/143 \\
AEB 5\,Hz & 0.104 & 0.230 & 0.072 & 0.015 & 82/143 \\
\bottomrule
\end{tabular}
\end{table}

\begin{table}[!htb]
\centering\scriptsize
\caption{Throughput saturation benchmark: logged rollout observations replayed continuously into 3 driver replicas (within the submission sandbox, video recording disabled) on a single shared NVIDIA RTX 6000 Ada GPU, with 2 concurrent sessions per replica and zero simulator rendering delay between consecutive RPC calls. Because eliminating simulator think-time exaggerates GPU contention, these values should be used only for relative comparisons across rows. All times are reported in seconds.}
\label{tab:app_load}
\begin{tabular}{lcccccccc}
\toprule
 & Sessions & Ticks & Mean & p50 & p90 & p99 & Per tick & Wall \\
\midrule
No AEB & 6 & 1071 & 0.294 & 0.001 & 1.457 & 1.512 & 0.322 & 64 \\
AEB 10\,Hz & 6 & 1071 & 0.317 & 0.169 & 0.890 & 1.125 & 0.346 & 71 \\
AEB 5\,Hz & 6 & 1071 & 0.306 & 0.113 & 1.167 & 1.365 & 0.333 & 67 \\
\bottomrule
\end{tabular}
\end{table}

In the back-to-back saturation benchmark (Table~\ref{tab:app_load}), enabling AEB increases the mean \texttt{Drive} call duration by 8\% at 10\,Hz and by 4\% at 5\,Hz relative to the policy without AEB. Note that the absolute latencies in this stress test do not reflect isolated model inference latency, as 6 sessions continuously contend for a single GPU with zero idle time between calls---unlike closed-loop simulation, where physics stepping and sensor rendering interleave with driver RPCs.

\FloatBarrier
\section{Sources of the reported data}
\label{app:sources}
All open-loop metrics, training curves, and visualizations in Figs.~\ref{fig:evo_left}--\ref{fig:final} are generated directly from W\&B runs \texttt{ccb412c9} and \texttt{03bdfc50} (project \texttt{starVLA\_drive}), both initialized from commit \texttt{b5c4446} with random seed 42. All closed-loop evaluation metrics are extracted from each run's \texttt{aggregate/results-summary.json}, per-step \texttt{metrics.parquet}, and \texttt{rollout.asl} logs. The model code packaged in the deployed container images differs from the training commit solely by configuration flags that remain disabled for this checkpoint (specifically, ego-history dropout and fallback endpoint target filtering).

\end{document}